\documentclass[preprint,12pt,authoryear]{elsarticle}

\usepackage{amssymb}
\usepackage{amsmath}

\usepackage{natbib}
\usepackage[utf8]{inputenc}
\usepackage{booktabs}
\usepackage{wrapfig}
\usepackage{caption}
\usepackage{subcaption}
\usepackage{float}
\usepackage{makecell}

\journal{University of Surrey}

\begin{document}

\begin{frontmatter}



\title{Efficient Deep Learning for Short-Term Solar Irradiance Time Series Forecasting: A Benchmark Study in Ho Chi Minh City}


\author{Tin Hoang} 

\ead{th01167@surrey.ac.uk}

\ead[url]{github.com/Tin-Hoang/solar-timeseries-forecasting}

\affiliation{organization={University of Surrey},
            city={Guildford},
            postcode={GU2 7XH}, 
            state={Surrey},
            country={UK}}

\begin{abstract}
Reliable forecasting of Global Horizontal Irradiance (GHI) is essential for mitigating the variability of solar energy in power grids. This study presents a comprehensive benchmark of ten deep learning architectures for short-term (1-hour ahead) GHI time series forecasting in Ho Chi Minh City, leveraging high-resolution NSRDB satellite data (2011–2020) to compare established baselines (LSTM, TCN) against emerging state-of-the-art architectures, including \textbf{Informer}, \textbf{iTransformer}, \textbf{TSMixer}, and the Selective State Space Model \textbf{Mamba}. Experimental results identify the \textbf{Transformer} as the superior architecture, achieving the highest predictive accuracy with an $R^2$ of 0.9696. The study further utilizes SHAP analysis to contrast the temporal reasoning of these architectures, revealing that Transformers exhibit a strong "recency bias" focused on immediate atmospheric conditions, whereas Mamba explicitly leverages 24-hour periodic dependencies to inform predictions. Furthermore, we demonstrate that \textbf{Knowledge Distillation} can compress the high-performance Transformer by 23.5\% while surprisingly reducing error (MAE: 23.78 W/m$^2$), offering a proven pathway for deploying sophisticated, low-latency forecasting on resource-constrained edge devices.
\end{abstract}



\begin{keyword}
Deep Learning \sep Time Series Forecasting \sep Renewable Energy Integration \sep AI for Sustainability \sep Explainable AI



\end{keyword}

\end{frontmatter}

\section{Introduction}

The global transition toward renewable energy sources represents one of the most crucial strategies to combat climate change and meet the accelerating global demand for energy. Among the array of renewable alternatives, solar energy stands out due to its abundance, ubiquity, and rapidly decreasing cost of deployment \cite{sengupta2018nsrdb}. Despite this promise, solar power generation remains inherently intermittent and variable, posing significant challenges to grid reliability, energy planning, and operational efficiency.

Accurate forecasting of solar irradiance, particularly Global Horizontal Irradiance (GHI), is essential for integrating solar power into modern energy systems. Reliable GHI prediction supports multiple critical domains, including grid load balancing, solar farm optimization, smart city infrastructure planning, and agricultural scheduling. For developing regions such as Ho Chi Minh City, Vietnam—characterized by high solar potential but limited forecasting infrastructure—such predictive capabilities are particularly transformative.

Recent advances in artificial intelligence (AI) offer new opportunities to overcome these forecasting challenges. Deep learning models can capture nonlinear dependencies and long-range temporal patterns in meteorological data, enabling the development of highly accurate solar radiation forecasting systems \cite{hochreiter1997lstm,vaswani2017attention,zhou2021informer}. By 2030, AI-driven technologies are projected to contribute up to USD 1.3 trillion in economic value to the energy sector while potentially reducing global greenhouse gas emissions by 5–10\% \cite{wef2025energyai}. However, the computational demands of AI raise an emerging concern—while AI can accelerate renewable energy adoption, it also risks amplifying energy consumption, creating a paradox between technological advancement and sustainability.

This research addresses these challenges by developing and evaluating computationally efficient deep learning models for short-term GHI forecasting using satellite-derived data from the National Solar Radiation Database (NSRDB). The study focuses specifically on Ho Chi Minh City, leveraging Himawari-7 satellite imagery from 2011–2020 to train, validate, and test multiple neural architectures, including both classical (LSTM, CNN-LSTM, MLP, and TCN) and state-of-the-art models (Transformer, Informer, TSMixer, iTransformer, and Mamba). The primary objectives of this work are to (1) assess the predictive performance of these architectures; (2) explore model interpretability through explainability techniques such as SHAP; (3) reduce computational footprint through quantization, pruning, and knowledge distillation; and (4) quantify the sustainability implications of accurate solar forecasting.

By contributing robust, interpretable, and energy-efficient forecasting methods, this research supports several United Nations Sustainable Development Goals (SDGs), including SDG 7 (Affordable and Clean Energy), SDG 9 (Industry, Innovation, and Infrastructure), and SDG 13 (Climate Action). It underscores the potential of AI as a catalyst for sustainable energy transition, particularly in fast-developing regions striving for energy resilience and environmental responsibility.
\section{Related Work}

Accurate forecasting of solar radiation, particularly Global Horizontal Irradiance (GHI), has long been a central challenge for solar energy integration. Traditional statistical and physical models such as ARIMA, persistence models, and physical clear-sky equations~\citep{sengupta2018nsrdb} lay the foundation for irradiance estimation but often fail to capture the non-linear spatiotemporal dependencies present in meteorological data. With the advent of deep learning, data-driven models have progressively replaced deterministic approaches, offering enhanced generalization and adaptability under complex weather conditions.

\subsection{Classical and Machine Learning Approaches}

Early research in GHI forecasting primarily leveraged linear regression and autoregressive models, relying on hand-crafted features derived from meteorological observations~\citep{voyant2017machine}. Subsequently, ensemble methods such as Random Forests and Gradient Boosting were introduced, showing improved performance through nonlinear pattern learning~\citep{abdel2020gradient}. However, their limited ability to handle temporal dependencies motivated the adoption of sequence modeling architectures.

\subsection{Recurrent and Convolutional Neural Networks}

Recurrent Neural Networks (RNNs), notably the Long Short-Term Memory (LSTM) network~\citep{hochreiter1997long}, became the dominant paradigm for solar forecasting due to their capacity to retain long-term dependencies. Studies have shown LSTMs outperform ARIMA and SVR models in both short-term and day-ahead forecasting~\citep{marquez2018comparison}. Hybrid architectures combining convolutional and recurrent layers, such as CNN-LSTM~\citep{omar2024lstmcnn}, exploit spatial features from satellite or sky imagery while learning temporal dynamics, achieving higher accuracy in cloudy-sky scenarios. More recent developments introduced Temporal Convolutional Networks (TCNs)~\citep{bai2018empirical}, which leverage dilated causal convolutions to model longer temporal dependencies efficiently while enabling parallel training—reducing computational time compared to sequential LSTM models.

\subsection{Transformer and Attention-based Models}

Transformers~\citep{vaswani2017attention} revolutionized sequential prediction by enabling direct long-range dependency modeling through self-attention, overcoming the limitations of vanishing gradients and fixed memory lengths in RNNs. In time-series forecasting, architectures such as Informer~\citep{zhou2021informer} introduced ProbSparse attention to reduce computational complexity and achieve efficient forecasting across long sequences. For solar forecasting tasks, Transformer variants have demonstrated strong capability in modeling global weather dynamics, outperforming traditional LSTM-based frameworks~\citep{lim2021temporal}. 

Following this trend, lightweight architectures such as TSMixer~\citep{chen2023tsmixer} and iTransformer~\citep{liu2024itransformer} have emerged, relying on MLP-like or inverted attention strategies for improved parameter efficiency and stability. Meanwhile, the recent Mamba model~\citep{gu2023mamba}, using Selective State Space Models (SSMs), achieves linear time complexity while maintaining competitive accuracy, positioning it as a promising successor to attention mechanisms in time-series modeling.

\subsection{AI for Sustainable Energy Forecasting}

As solar energy systems expand globally, AI-based forecasting has become a cornerstone of grid management and smart city energy planning. Studies integrating AI in renewable forecasting show potential to reduce grid uncertainty and fossil fuel dependence~\citep{wef2025}. However, growing emphasis is placed on the sustainability of AI itself; training large models can incur substantial carbon footprints~\citep{strubell2019energy}. To mitigate this, modern research explores model compression, quantization, and knowledge distillation to optimize energy efficiency without compromising performance~\citep{hinton2015distilling}. 

This study contributes to this literature by comparing classic and state-of-the-art deep learning architectures under a sustainability lens. It emphasizes both performance and computational footprint, applying model compression and explainability techniques to ensure practical, eco-efficient deployment of GHI prediction systems for developing urban contexts such as Ho Chi Minh City.
\section{Data Acquisition and Preparation}
\label{sec:data_prep}

\begin{wrapfigure}{r}{0.5\textwidth}
    \centering
    \includegraphics[width=0.5\textwidth]{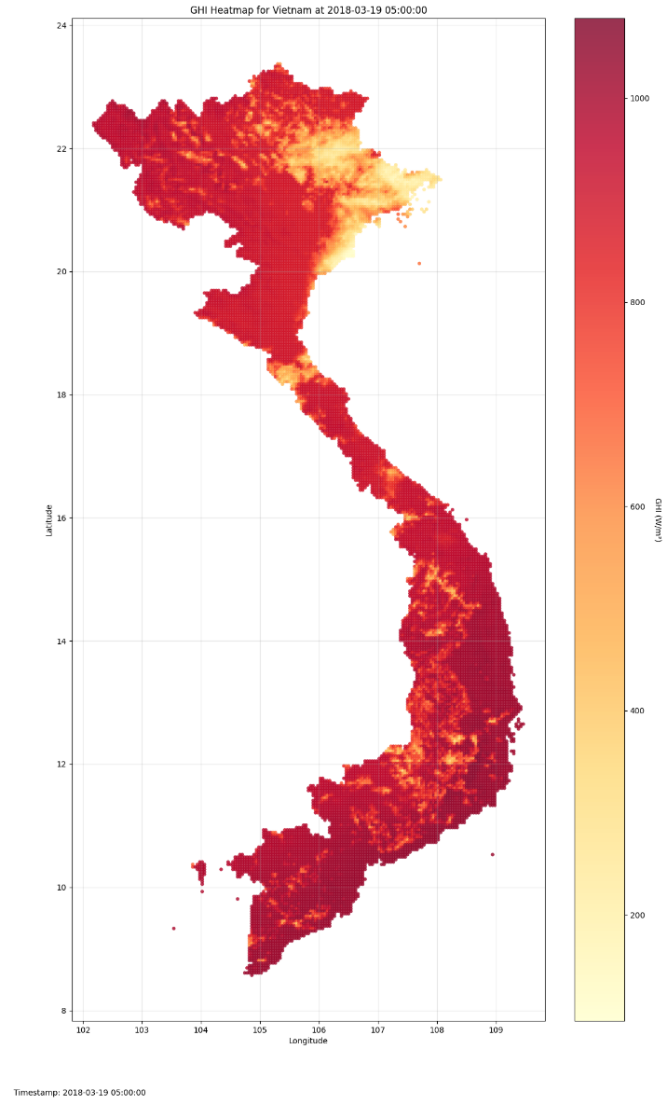}
    \caption{Sample heatmap of GHI across Vietnam at 2018-03-19 12:00 VNT, showcasing regional solar radiation differences.}
    \label{fig:vietnam-heatmap}
\end{wrapfigure}

This section details the provenance of the solar irradiance data and the rigorous pre-processing pipeline implemented to transform raw satellite observations into structured tensor inputs suitable for deep learning. The pipeline encompasses data cleaning, feature engineering, and sliding-window generation, ensuring that the models receive high-quality, physically consistent signals.

\subsection{Data Source Description}

The primary dataset for this study is sourced from the \textbf{National Solar Radiation Database (NSRDB)} \cite{sengupta2018nsrdb}. This source was selected for its high fidelity and extensive spatial coverage, derived from the Himawari-7 satellite's Physical Solar Model Version 3 (PSM v3). Unlike ground-station measurements, which may suffer from sensor downtime, satellite-derived data ensures a continuous, gap-free time series, which is critical for training recurrent neural networks without introducing interpolation artifacts.

The dataset spans the years \textbf{2011–2020} with a temporal resolution of \textbf{30 minutes} and a spatial resolution of approximately 4 km. It targets 105 grid cells encompassing the Ho Chi Minh City region, a tropical locale characterized by rapid cloud formation and high solar variability.

\begin{wrapfigure}{r}{0.5\textwidth}
    \centering
    \includegraphics[width=0.5\textwidth]{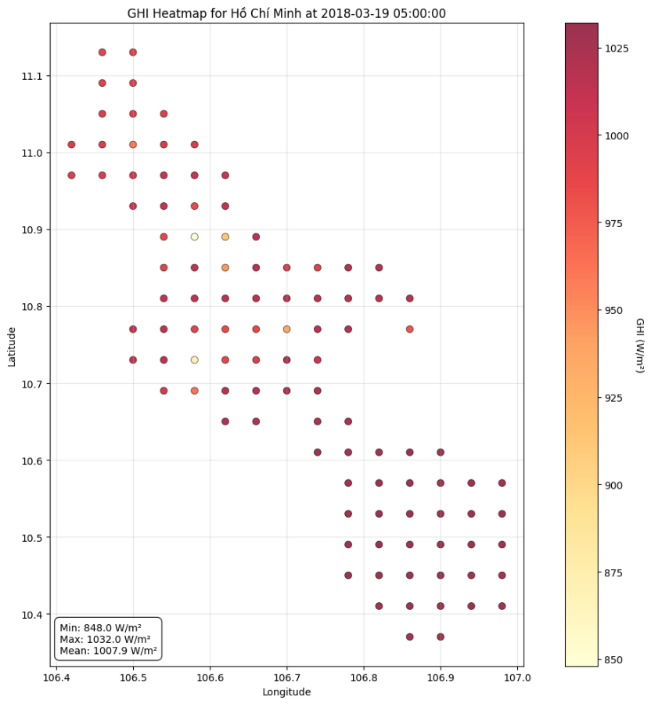}
    \caption{Sample heatmap of GHI for Ho Chi Minh City on 2018-03-19, illustrating spatial variability across 105 grid cells.}
    \label{fig:hcm-heatmap}
\end{wrapfigure}

\subsection{Exploratory Data Analysis (EDA)}
Prior to modeling, Exploratory Data Analysis was conducted to understand the statistical properties and quality of the data. Understanding the underlying distribution of the target variable (GHI) is crucial for identifying anomalies and verifying physical consistency. The data exhibits a bimodal distribution, peaking at noon and dropping to zero at night, which necessitates specific handling strategies such as nighttime masking.

Additionally, spatial analysis (Figure \ref{fig:hcm-heatmap}) confirmed that micro-climate variations exist even within the Ho Chi Minh City region. This observation justifies the inclusion of location-specific static features (latitude, longitude, elevation), allowing the models to generalize across different grid cells.

\subsection{Data Pre-processing Strategy}
Raw satellite data requires significant cleaning and transformation to be usable by gradient-based learning algorithms. The following steps were implemented to ensure data quality and model stability.

\subsubsection{Data Conversion and Cleaning}
The raw NSRDB data, originally stored in HDF5 format with integer encoding for storage optimization, was converted back to physical floating-point representations using NSRDB-specific scaling factors. Deep learning models require high-precision continuous inputs to capture subtle gradients in meteorological changes; integer approximations would introduce quantization noise capable of degrading forecast accuracy.

\subsubsection{Nighttime Masking}
\begin{wrapfigure}{r}{0.6\textwidth}
    \centering
    \includegraphics[width=0.58\textwidth]{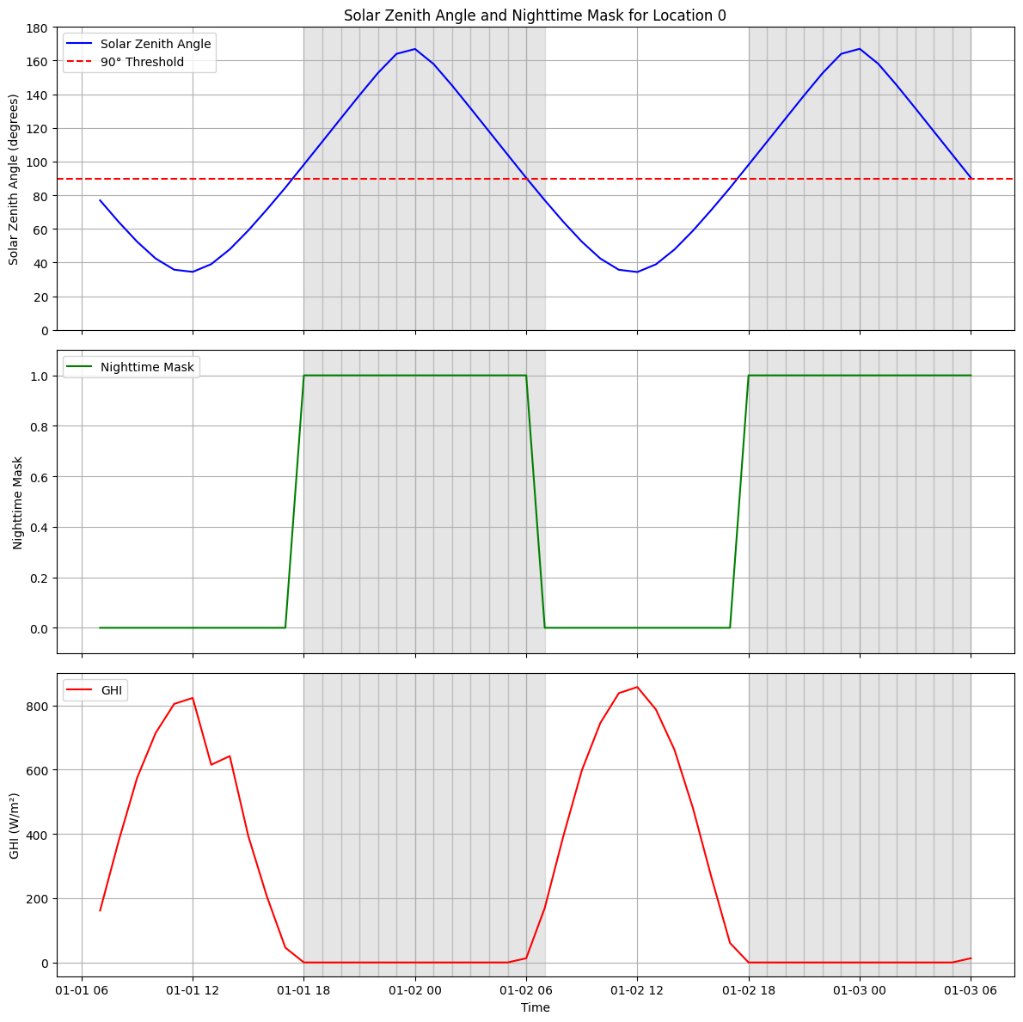}
    \caption{Visualization of nighttime mask calculated by Solar Zenith Angle. The nighttime is 1 when Zenith angle $\geq 90^\circ$.}
    \label{fig:nighttime-map}
\end{wrapfigure}

Solar irradiance is naturally zero during the night. Including these trivial zero-values in the training process can bias the model, causing it to over-prioritize the trivial task of predicting nighttime zeros rather than learning the complex daytime dynamics. To address this, a binary \texttt{nighttime\_mask} was generated based on the solar zenith angle ($z$):

\begin{equation}
    \text{Mask} = 
    \begin{cases} 
    1 & \text{if } z \geq 90^\circ \text{ (Night)} \\
    0 & \text{if } z < 90^\circ \text{ (Day)}
    \end{cases}
\end{equation}

This mask enables the loss function to selectively focus on daytime intervals where the forecasting challenge lies.

\subsubsection{Clearsky GHI Computation}
To provide the models with a physical baseline, the theoretical "Clear-sky GHI"—the maximum possible irradiance in the absence of clouds—was computed. This was derived using the solar zenith angle ($z$) and the solar constant ($I_0 = 1366.1 \, W/m^2$) with an approximate atmospheric transmittance of 0.7:

\begin{equation}
    \text{GHI}_{\text{clearsky}} = \max\left(0, I_0 \cdot \cos(z) \cdot 0.7\right)
\label{eq:clearsky_ghi}
\end{equation}

\begin{figure}[h]
    \centering
    \includegraphics[width=0.85\textwidth]{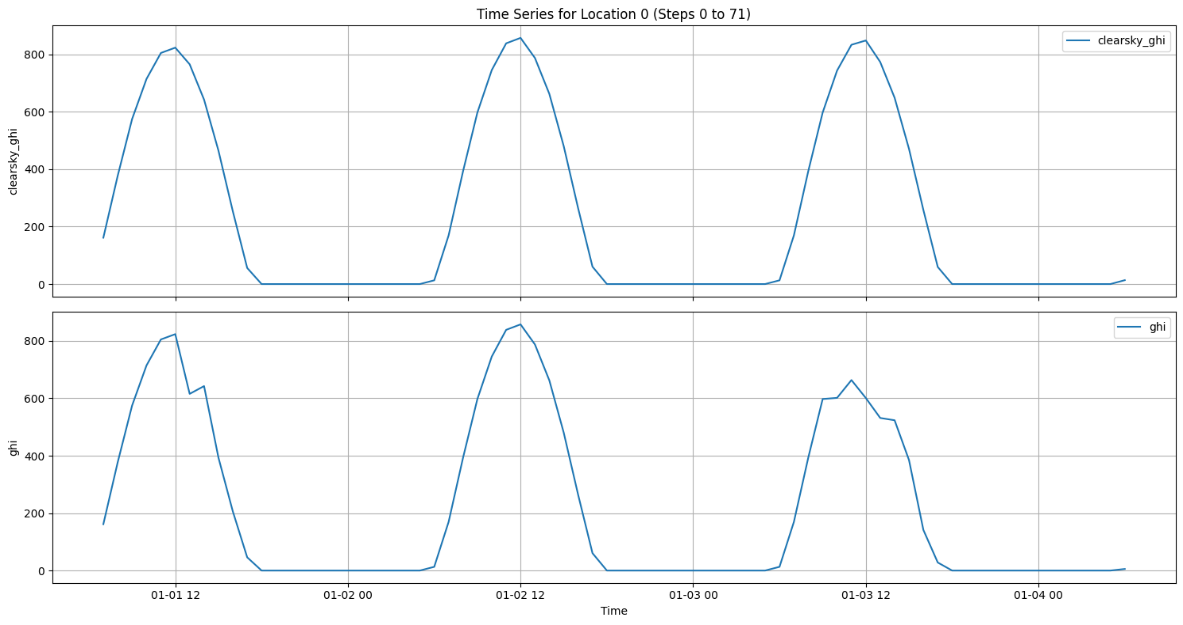}
    \caption{Visualization of Clearsky GHI vs. Actual GHI. Clearsky GHI represents the theoretical maximum in ideal conditions.}
    \label{fig:clearsky_ghi}
\end{figure}

Providing this theoretical maximum helps the model distinguish between reduced irradiance caused by the sun's position (geometric drop) versus reductions caused by cloud cover (meteorological drop), significantly simplifying the learning task.

\subsection{Feature Engineering and Selection}

\subsubsection{Cyclical Time Encoding}
Standard integer representations of time (e.g., Hour 0 following Hour 23) introduce numerical discontinuities that can confuse neural networks. To preserve the temporal continuity of the daily cycle, temporal variables (hour, day, month) were transformed into continuous cyclical features using sine and cosine functions:

\begin{equation}
    x_{sin} = \sin\left(\frac{2\pi x}{T}\right), \quad x_{cos} = \cos\left(\frac{2\pi x}{T}\right)
\end{equation}

where $T$ is the period (e.g., 24 for hours). This encoding ensures that "23:00" and "00:00" are numerically close in the feature space, accurately reflecting their temporal proximity.

\begin{figure}[H]
    \centering
    \includegraphics[width=0.85\textwidth]{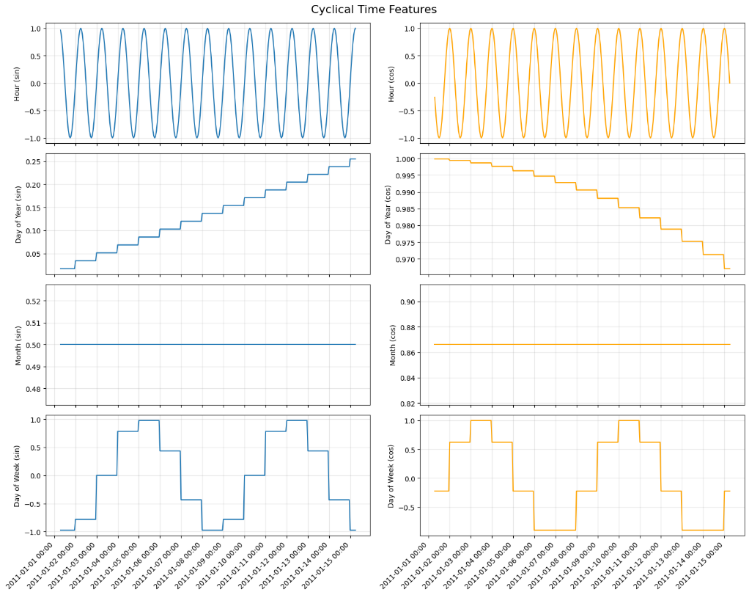}
    \caption{Engineered cyclical time features derived from timestamps.}
    \label{fig:time-freq}
\end{figure}

\subsubsection{Input Feature Selection}
While the raw NSRDB dataset contains 26 variables (see Appendix A for the full data dictionary), a subset of features most relevant to radiative transfer physics was selected to mitigate the "curse of dimensionality." The final input vector $x_t$ comprises:

\begin{itemize}
    \item \textbf{Target:} Global Horizontal Irradiance (GHI).
    \item \textbf{Meteorological:} Air Temperature, Relative Humidity, Wind Speed, Total Precipitable Water, Cloud Type.
    \item \textbf{Solar Geometry:} Solar Zenith Angle, Azimuth Angle, Clearsky GHI.
    \item \textbf{Engineered:} Cyclical time features (sin/cos for hour, day, month), Nighttime Mask.
    \item \textbf{Static:} Latitude, Longitude, Elevation.
\end{itemize}

\subsubsection{Normalization}
To facilitate efficient gradient-based optimization, all dynamic features were normalized. \textbf{Min-Max Scaling} was applied to transform meteorological variables into the $[0, 1]$ range. Deep learning optimizers like Adam converge faster and more stably when input features share a similar scale; without normalization, variables with large magnitudes (e.g., GHI $\approx$ 1000) would dominate the gradient updates over smaller variables (e.g., Humidity $\approx$ 0.8). Scalers were fitted \textit{exclusively} on the training set (2011–2018) to prevent data leakage.

\subsection{Input Generation (Sliding Window Strategy)}
To formulate the time-series forecasting task as a supervised learning problem, a sliding window approach was employed. A \textbf{Lookback Window ($L=24$)} representing the past 24 hours (48 steps at 30-min intervals) was defined to predict the \textbf{Forecast Horizon ($H=1$)} (1 hour ahead).

A 24-hour window captures exactly one full diurnal cycle. This enables the model to observe the previous day's pattern at the same timestamp, allowing it to model persistence (e.g., historical sunny conditions) and immediate trends (e.g., increasing cloud cover over the last 3 hours).

\subsection{Data Splitting}
The dataset was split chronologically to simulate a realistic operational scenario:
\begin{itemize}
    \item \textbf{Training (2011–2018):} 8 years for learning patterns.
    \item \textbf{Validation (2019):} 1 year for hyperparameter tuning and early stopping.
    \item \textbf{Testing (2020):} 1 year for final unbiased evaluation.
\end{itemize}
Chronological splitting is strictly preferred over random shuffling for time-series data to avoid "look-ahead bias," where the model inadvertently learns from future samples.

\newpage
\section{Proposed Framework}
\label{sec:methodology}

Having established a robust data processing pipeline, this section details the forecasting framework designed to map historical meteorological observations to future irradiance levels. The methodology encompasses the mathematical formulation of the learning task, the architectural specifications of the baseline and advanced models, and the training strategies employed to ensure convergence and generalization.

\subsection{Problem Formulation}
The Global Horizontal Irradiance (GHI) forecasting task is formulated as a supervised time-series regression problem. Given a historical sequence of input vectors $X_{t-L}, \dots, X_t$, where $L=24$ represents a lookback window covering one complete diurnal cycle, and each $x_t \in \mathbb{R}^F$ is a vector of $F$ features, the objective is to learn a mapping function $f(\cdot)$ that predicts the GHI value $y_{t+H}$ for a horizon $H=1$ (one hour ahead):

\begin{equation}
    \hat{y}_{t+H} = f(X_{t-L:t}; \theta)
\end{equation}

where $\theta$ represents the learnable parameters of the neural network. This sequence-to-one formulation allows the model to directly optimize for specific operational planning horizons, which is critical for grid stability management.

\subsection{Basic Architectures}
To establish a rigorous performance benchmark, several fundamental neural network architectures were implemented. These models represent distinct approaches to temporal modeling, ranging from simple static mappings to recurrent mechanisms.

\begin{itemize}
    \item \textbf{Multilayer Perceptron (MLP)} \cite{rosenblatt1958perceptron}: As a foundational baseline, the MLP processes the flattened temporal window as a single static vector. This architecture tests the hypothesis that complex temporal dependencies might be secondary to immediate feature interactions, serving as a lower-bound for computational complexity.
    
    \item \textbf{Long Short-Term Memory (LSTM)} \cite{hochreiter1997long}: Standard Recurrent Neural Networks (RNNs) often struggle with vanishing gradients over long sequences. The LSTM addresses this through gating mechanisms that explicitly regulate information flow, allowing the model to retain relevant context from the start of the 24-hour window while discarding noise. Our implementation utilizes 2 stacked LSTM layers with 128 hidden units and a dropout rate of 0.2 to prevent overfitting.

    \item \textbf{1D Convolutional Neural Network (1D-CNN)} \cite{kiranyaz20211d}: While RNNs process data sequentially, CNNs process temporal data in parallel using sliding filters. This architecture is adept at extracting local invariant features—such as short-term cloud movements—and is computationally more efficient than recurrent architectures due to parallelizability.

    \item \textbf{CNN-LSTM} \cite{shi2015convlstm}: This hybrid architecture leverages the strengths of both preceding models. Convolutional layers first extract high-level local features from the input sequence, which are then fed into LSTM layers to model temporal dynamics. This hierarchical approach aims to capture both short-term weather fluctuations and daily solar periodicities.

    \item \textbf{Temporal Convolutional Network (TCN)} \cite{bai2018empirical}: To capture long-range dependencies without the sequential bottleneck of RNNs, the TCN employs dilated causal convolutions. The dilation factor increases exponentially with depth, creating a large receptive field that allows the model to "see" the entire history of the input window simultaneously.
\end{itemize}

\subsection{Advanced Architectures}
Beyond traditional baselines, this study explores state-of-the-art architectures designed to handle complex dependencies and improve computational scaling.

\begin{itemize}
    \item \textbf{Transformer} \cite{vaswani2017attention}: Relying entirely on self-attention mechanisms, the Transformer captures global dependencies regardless of temporal distance. By weighing the importance of every time step against every other, it can identify non-local correlations—such as the relationship between yesterday's noon irradiance and today's—that RNNs might miss. Our configuration features 2 encoder layers with 4 attention heads ($d_{model}=128$).

    \item \textbf{Informer} \cite{zhou2021informer}: Standard Transformers scale quadratically ($O(L^2)$) with sequence length, which limits efficiency. The Informer introduces a \textit{ProbSparse} attention mechanism to select only the most dominant queries, reducing complexity to $O(L \log L)$. This architecture is included to evaluate the trade-off between sparse attention efficiency and forecasting accuracy in the context of solar data.

    \item \textbf{TSMixer} \cite{chen2023tsmixer}: Challenging the prevailing dominance of attention mechanisms, TSMixer utilizes an all-MLP architecture that alternately mixes features across the time and feature dimensions. This design tests whether modern MLP-based structures can achieve competitive performance with significantly lower latency and computational overhead than Transformers.

    \item \textbf{iTransformer} \cite{liu2023itransformer}: Unlike standard Transformers that apply attention across time, the iTransformer applies attention and feed-forward networks across the \textit{feature} dimension. This inverted structure effectively captures multivariate correlations, prioritizing the relationships between meteorological variables (e.g., humidity vs. temperature) over purely temporal interactions.

    \item \textbf{Mamba} \cite{gu2023mamba}: Representing the frontier of sequence modeling, Mamba is based on Selective State Space Models (SSMs). It achieves linear scaling with sequence length ($O(L)$) and integrates a selection mechanism into the state space transition. This allows the model to selectively propagate or forget information based on the input content, theoretically combining the training efficiency of CNNs with the inference power of RNNs.
\end{itemize}

\subsection{Training Strategy}
To ensure a fair comparison, all models were trained using a unified pipeline:
\begin{itemize}
    \item \textbf{Optimization:} We used the \texttt{AdamW} optimizer with a learning rate of $1 \times 10^{-4}$ and a weight decay of $0.01$. A \texttt{ReduceLROnPlateau} scheduler adjusted the learning rate based on validation loss.
    \item \textbf{Loss Function:} Mean Squared Error (MSE) was utilized as the objective function.
    \item \textbf{Training Regime:} Models were trained for a maximum of 30 epochs with a batch size of 8192 to maximize GPU throughput. Early stopping with a patience of 5 epochs was implemented to prevent overfitting.
\end{itemize}

\subsection{Evaluation Protocols}
To thoroughly assess the predictive performance and operational viability of the proposed models, a comprehensive set of statistical metrics was employed.

\subsubsection{Accuracy Metrics}
Given the ground truth values $y_i$ and predicted values $\hat{y}_i$ for $N$ samples, the following metrics were calculated:

\begin{itemize}
    \item \textbf{Mean Squared Error (MSE):}
    \begin{equation}
        \text{MSE} = \frac{1}{N} \sum_{i=1}^{N} (y_i - \hat{y}_i)^2
    \end{equation}
    MSE serves as the primary loss function during training. By squaring the residuals, it penalizes larger errors more heavily, which is critical for grid stability where large, unexpected drops in solar generation can be costly.

    \item \textbf{Root Mean Squared Error (RMSE):}
    \begin{equation}
        \text{RMSE} = \sqrt{\frac{1}{N} \sum_{i=1}^{N} (y_i - \hat{y}_i)^2}
    \end{equation}
    RMSE retains the sensitivity to large errors but is expressed in the same units as the target variable ($W/m^2$), facilitating easier interpretation of the error magnitude.

    \item \textbf{Mean Absolute Error (MAE):}
    \begin{equation}
        \text{MAE} = \frac{1}{N} \sum_{i=1}^{N} |y_i - \hat{y}_i|
    \end{equation}
    MAE provides a linear representation of the average error. It is less sensitive to outliers than MSE/RMSE and offers a robust indicator of typical model performance across varying weather conditions.

    \item \textbf{Coefficient of Determination ($R^2$):}
    \begin{equation}
        R^2 = 1 - \frac{\sum_{i=1}^{N} (y_i - \hat{y}_i)^2}{\sum_{i=1}^{N} (y_i - \bar{y})^2}
    \end{equation}
    where $\bar{y}$ is the mean of the observed data. $R^2$ measures the proportion of variance in the solar irradiance explained by the model, indicating how well the model captures the complex variability of GHI compared to a simple mean baseline.

    \item \textbf{Mean Absolute Scaled Error (MASE) \cite{hyndman2006another}:}
    \begin{equation}
        \text{MASE} = \frac{\text{MAE}}{\frac{1}{N-1} \sum_{i=2}^{N} |y_i - y_{i-1}|}
    \end{equation}
    MASE is a scale-independent metric that compares the model's error to that of a naïve persistence forecast (where the prediction equals the previous observed value). A MASE value $<1$ indicates the model outperforms the naïve baseline, validating its practical utility.
\end{itemize}
\subsubsection{Daytime vs. Nighttime Evaluation}
Solar irradiance data contains frequent zero values during nighttime hours. Including these in aggregate metrics can artificially deflate error rates and inflate accuracy scores. To ensure a rigorous evaluation, we computed metrics separately for:
\begin{itemize}
    \item \textbf{Daytime ($z < 90^\circ$):} The critical period for energy generation.
    \item \textbf{Nighttime ($z \geq 90^\circ$):} Used primarily to verify that models correctly predict zero irradiance without "hallucinating" solar output.
\end{itemize}

\subsubsection{Computational Efficiency:}
To assess suitability for real-time deployment, we measured inference efficiency using two metrics:
\begin{itemize}
    \item \textbf{Latency:} The average time required to process a single sample ($\mu s/\text{sample}$).
    \item \textbf{Throughput:} The number of samples processed per second.
\end{itemize}

\subsection{Interpretability and Efficiency Framework}
To bridge the gap between black-box predictions and actionable grid insights, this study integrates specific mechanisms for explainability and sustainability.

\subsubsection{SHAP (SHapley Additive exPlanations)}
To provide local interpretability, we utilized the SHAP framework \cite{lundberg2017unified}, specifically the \texttt{GradientExplainer}, which leverages the differentiability of deep learning models to approximate Shapley values efficiently.
The analysis focused exclusively on daytime timesteps, as nighttime GHI is trivially zero. The explainer was initialized with a background dataset of 20 samples from the training set to establish a baseline. SHAP values were then computed for 400 test samples, quantifying the contribution of each feature—across the 24-hour lookback window—to the final GHI prediction.

\subsubsection{Sustainability Optimization}
Aligning with UN Sustainable Development Goals (SDG 7 and 13), we explored techniques to minimize the computational footprint of our models.

\paragraph{Quantization:}
We applied dynamic quantization using the Open Neural Network Exchange (ONNX) runtime \cite{bai2019onnx}. This process reduces the precision of model weights and activations from 32-bit floating-point (FP32) to 8-bit integers (INT8). By compressing the numerical representation, we significantly reduced memory bandwidth requirements and inference latency, making the models more suitable for resource-constrained edge devices in solar farms.

\paragraph{Structured Pruning:}
To remove redundant parameters, we employed structured pruning \cite{han2015deep}. This technique systematically removes entire channels or neurons with the smallest $L_1$-norms, rather than individual weights. This results in a denser, more efficient architecture that maintains hardware friendliness compared to unstructured sparsity.

\paragraph{Knowledge Distillation:}
We utilized Knowledge Distillation (KD) \cite{hinton2015distilling} to transfer the generalization capabilities of a heavy, high-performance "teacher" model (e.g., Transformer) to a compact "student" architecture (e.g., MLP). The student is trained to minimize the divergence between its predictions and the teacher's soft targets, effectively learning a compressed representation of the teacher's knowledge.
\newpage
\section{Results}
\label{sec:results}

In this section, we evaluate the performance of the implemented deep learning models for 1-hour ahead Global Horizontal Irradiance (GHI) forecasting. The evaluation is divided into two categories: Basic Models (LSTM, CNN-LSTM, MLP, 1D-CNN, TCN) and Advanced Models (Transformer, Informer, TSMixer, iTransformer, Mamba). We analyze their predictive accuracy using metrics such as Mean Squared Error (MSE), Root Mean Squared Error (RMSE), Mean Absolute Error (MAE), Mean Absolute Scaled Error (MASE), and the Coefficient of Determination ($R^2$). Additionally, we assess computational efficiency through inference throughput (samples/sec).

\subsection{Basic Models Performance}
\begin{figure}[H]
    \centering
    \includegraphics[width=\linewidth]{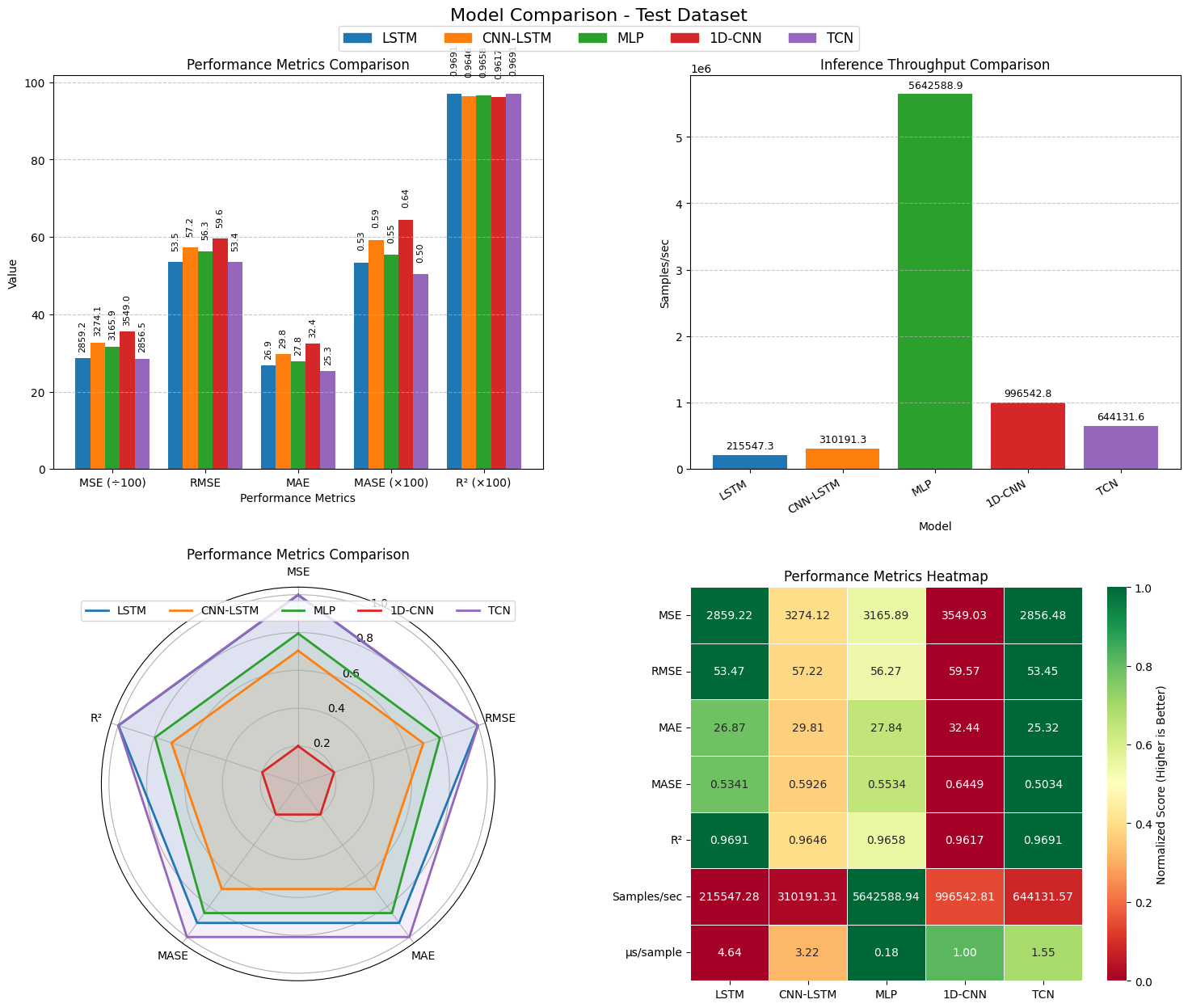}
    \caption{Performance evaluation of 5 basic models: (a) Metrics Comparison, (b) Inference Throughput, (c) Radar Chart of standardized metrics, and (d) Heatmap of raw performance values.}
    \label{fig:basic_results}
\end{figure}

The performance of the five basic models was evaluated on the test dataset. Figure \ref{fig:basic_results} illustrates the comparative metrics and inference throughput for these architectures. 

Among the basic models, the \textbf{Temporal Convolutional Network (TCN)} achieved the highest accuracy, recording the lowest error rates across most metrics (MSE: 2856.48, MAE: 25.32) and a high $R^2$ of 0.9691. The \textbf{LSTM} model followed closely with competitive accuracy (MSE: 2859.22) but demonstrated significantly lower inference speed compared to the convolutional and dense architectures. Conversely, the \textbf{Multi-Layer Perceptron (MLP)} demonstrated exceptional computational efficiency, achieving the highest throughput of approximately 5.6 million samples/sec, making it a strong candidate for resource-constrained environments despite slightly lower predictive accuracy compared to TCN and LSTM.

\subsection{Advanced Models Performance}
\begin{figure}[H]
    \centering
    \includegraphics[width=\linewidth]{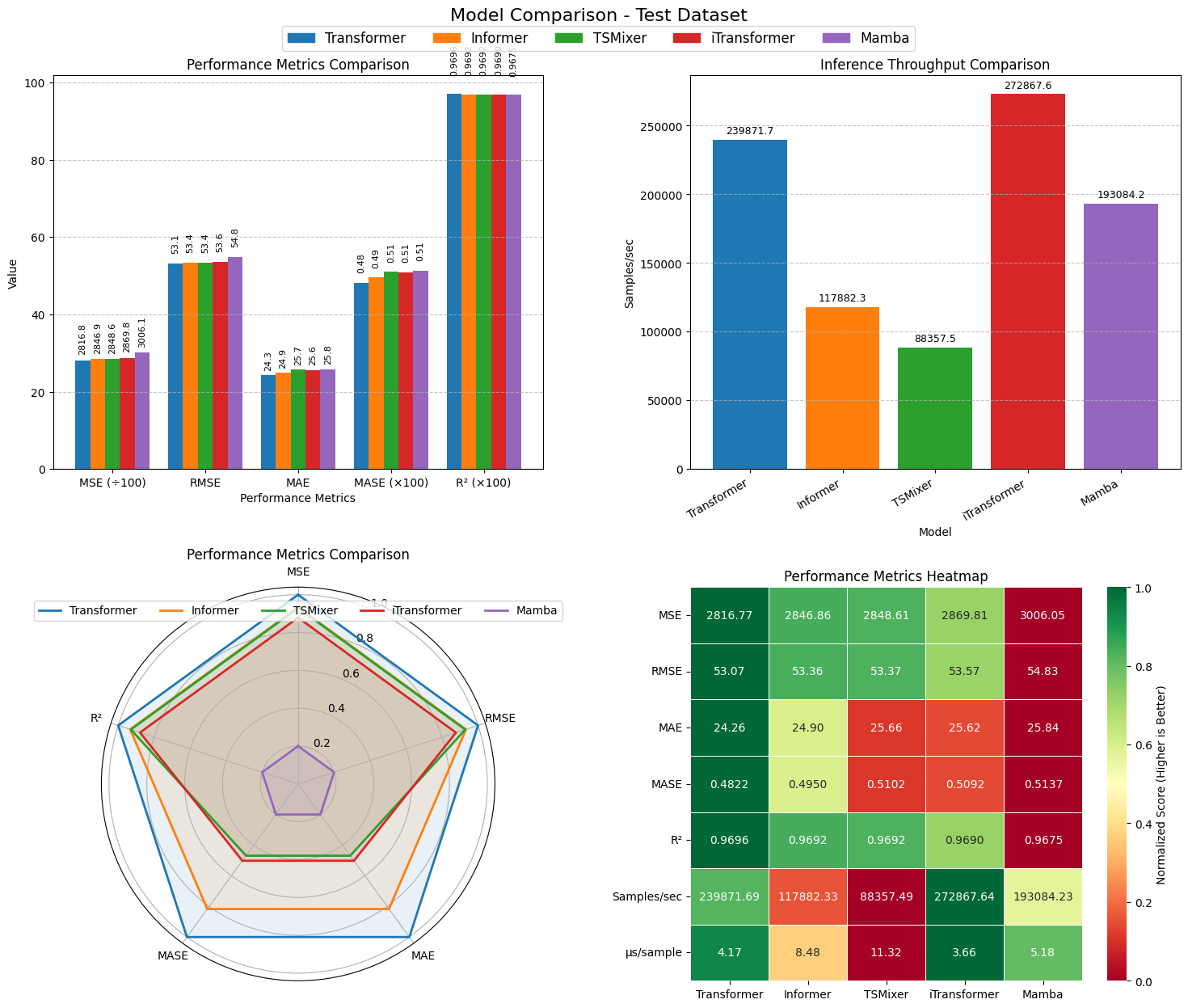}
    \caption{Performance evaluation of 5 advanced models: (a) Metrics Comparison, (b) Inference Throughput, (c) Radar Chart of standardized metrics, and (d) Heatmap of raw performance values.}
    \label{fig:advanced_results}
\end{figure}

The advanced architectures, primarily transformer-based and state-space models, were evaluated under the same conditions. Figure \ref{fig:advanced_results} presents the detailed results for these models.

The \textbf{Transformer} model emerged as the top performer among all evaluated architectures, achieving the best overall accuracy with an MSE of 2816.77 and an MAE of 24.26. It effectively captured complex temporal dependencies, outperforming newer architectures like Informer and TSMixer by a small margin. The \textbf{Mamba} model, despite its theoretical advantages in sequence modeling, showed the weakest performance in this specific GHI forecasting task (MSE: 3006.05). However, it maintained a reasonable inference speed. \textbf{iTransformer} demonstrated robust throughput (approx. 272k samples/sec), striking a good balance between complexity and speed.

\subsection{Comprehensive Model Comparison}
A consolidated comparison of all 10 models is presented in Table \ref{tab:model_comparison}. The results highlight a trade-off between accuracy and computational cost. While the Transformer provides the most precise forecasts, the TCN offers a highly competitive alternative with lower structural complexity. For real-time applications requiring ultra-low latency, the MLP remains unmatched in speed.

\begin{table}[htbp]
\centering
\caption{Comprehensive Performance Comparison of All 10 Models}
\label{tab:model_comparison}
\resizebox{\textwidth}{!}{%
\begin{tabular}{lcccccc}
\hline
\textbf{Model} & \textbf{MSE} & \textbf{RMSE} & \textbf{MAE} & \textbf{MASE} & \textbf{$R^2$} & \textbf{Speed (samples/sec)} \\ \hline
\multicolumn{7}{c}{\textbf{Basic Models}} \\
LSTM & 2859.22 & 53.47 & 26.87 & 0.5341 & 0.9691 & 215,547 \\
CNN-LSTM & 3274.12 & 57.22 & 29.81 & 0.5926 & 0.9646 & 310,191 \\
MLP & 3165.89 & 56.27 & 27.84 & 0.5534 & 0.9658 & 5,642,588 \\
1D-CNN & 3549.03 & 59.57 & 32.44 & 0.6449 & 0.9617 & 996,542 \\
\textbf{TCN} & \textbf{2856.48} & \textbf{53.45} & \textbf{25.32} & \textbf{0.5034} & \textbf{0.9691} & 644,131 \\ \hline
\multicolumn{7}{c}{\textbf{Advanced Models}} \\
\textbf{Transformer} & \textbf{2816.77} & \textbf{53.07} & \textbf{24.26} & \textbf{0.4822} & \textbf{0.9696} & 239,871 \\
Informer & 2846.86 & 53.36 & 24.90 & 0.4950 & 0.9692 & 117,882 \\
TSMixer & 2848.61 & 53.37 & 25.66 & 0.5102 & 0.9692 & 88,357 \\
iTransformer & 2869.81 & 53.57 & 25.62 & 0.5092 & 0.9690 & 272,867 \\
Mamba & 3006.05 & 54.83 & 25.84 & 0.5137 & 0.9675 & 193,084 \\ \hline
\end{tabular}%
}
\end{table}

Overall, the Transformer outperforms all other models, reducing the MSE by approximately 1.4\% compared to the best basic model (TCN). However, the basic TCN remains a robust choice, effectively matching the performance of more complex advanced architectures like TSMixer and Informer while offering significantly higher throughput.

\subsection{Models’ Parameters and Training Times}
To ensure a fair comparison between architectures of varying complexity, all models were designed to have a comparable total number of parameters. Table \ref{tab:model_params} presents the detailed parameter counts and training durations for all ten models implemented in this study.

\begin{table}[H]
    \centering
    \caption{Comparison of Parameters and Training Times for All 10 Models}
    \label{tab:model_params}
    \begin{tabular}{l c c}
        \toprule
        \textbf{Model} & \textbf{\# Parameters} & \textbf{Training Time} \\
        \midrule
        \multicolumn{3}{c}{\textit{Basic Models}} \\
        LSTM & 221,441 & 1h 3m 11s \\
        CNN-LSTM & 306,049 & 35m 41s \\
        MLP & 425,793 & 28m 46s \\
        1D-CNN & 176,097 & 46m 16s \\
        TCN & 82,273 & 47m 52s \\
        \midrule
        \multicolumn{3}{c}{\textit{Advanced Models}} \\
        Transformer & 280,449 & 49m 39s \\
        Informer & 334,977 & 1h 18m 34s \\
        TSMixer & 109,823 & 54m 50s \\
        iTransformer & 271,106 & 1h 19m 6s \\
        Mamba & 250,049 & 1h 6m 19s \\
        \bottomrule
    \end{tabular}
\end{table}

\subsection{Temporal Performance Analysis}
To provide a deeper insight into model robustness, we segregated performance into daytime and nighttime conditions. Since GHI is zero at night, this distinction prevents zero-value predictions from artificially inflating overall accuracy metrics. The detailed performance breakdowns are presented in Figure \ref{fig:basic_daytime_nighttime} for Basic Models and Figure \ref{fig:advanced_daytime_nighttime} for Advanced Models.

\subsubsection{Daytime Performance}
During active solar periods, the models must capture high-variance irradiance fluctuations. As shown in Figure \ref{fig:advanced_daytime_nighttime}, the \textbf{Transformer} model demonstrates superior performance across all metrics, achieving the lowest errors (MSE: 5653.55, RMSE: 75.19) and the highest $R^2$ of 0.9313. Conversely, \textbf{Mamba} performs the worst among advanced models during the day (MSE: 6025.45), indicating a struggle to capture complex daytime dynamics compared to the attention-based architectures.

\begin{figure}[H]
    \centering
    \includegraphics[width=0.95\linewidth]{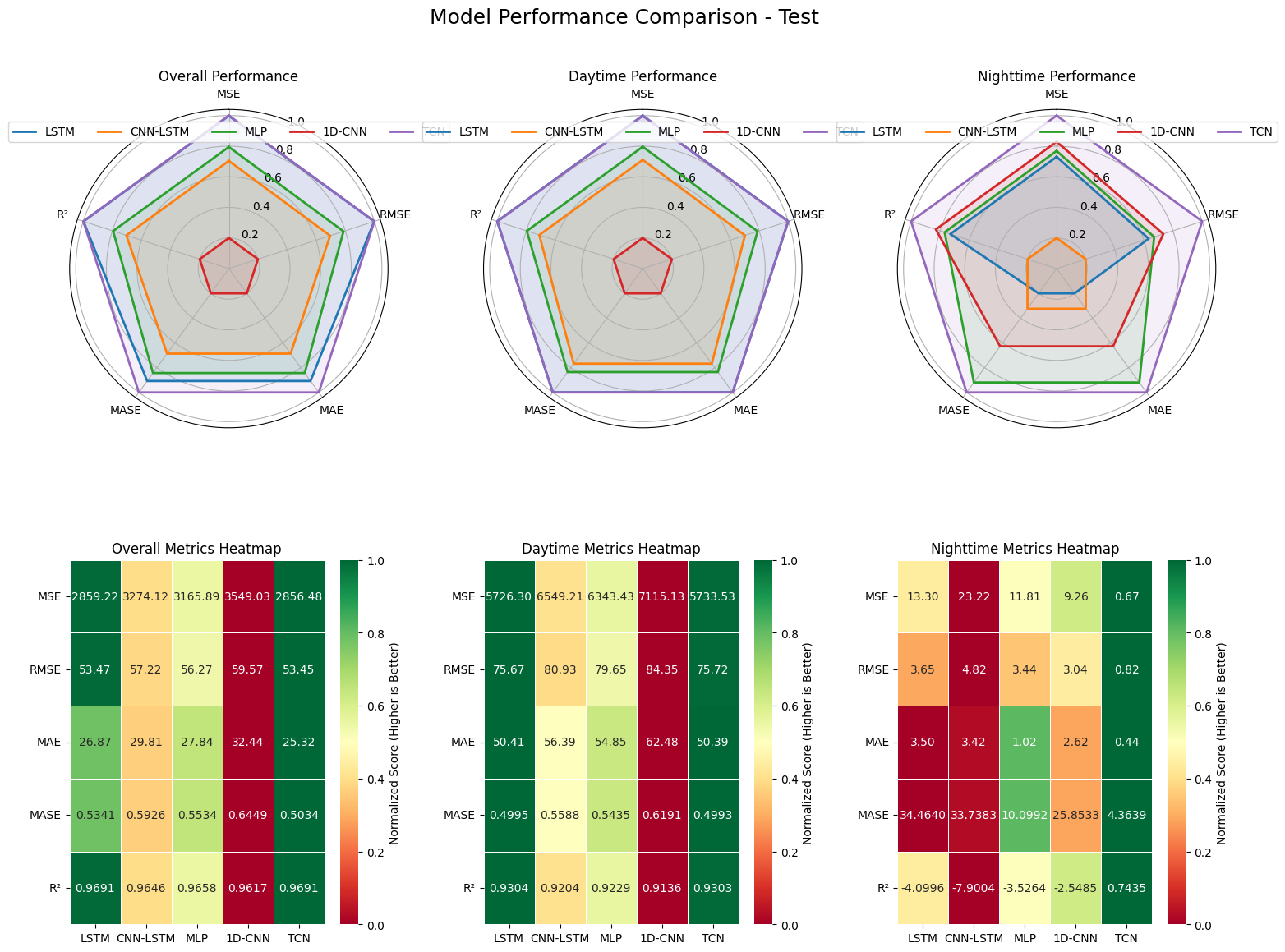}
    \caption{Comparing performance metrics of Basic models across Overall, Daytime, and Nighttime conditions.}
    \label{fig:basic_daytime_nighttime}
\end{figure}

Among the basic models (Figure \ref{fig:basic_daytime_nighttime}), \textbf{LSTM} and \textbf{TCN} are the top performers. LSTM slightly outperforms TCN in MSE (5726.30 vs. 5733.53) and $R^2$. In contrast, \textbf{1D-CNN} is the weakest basic performer, lagging significantly in accuracy. This comparison highlights that while TCN and Transformer effectively handle solar variability, 1D-CNN and Mamba are less reliable for peak irradiance forecasting.

\subsubsection{Nighttime Performance}
\begin{figure}[H]
    \centering
    \includegraphics[width=0.95\linewidth]{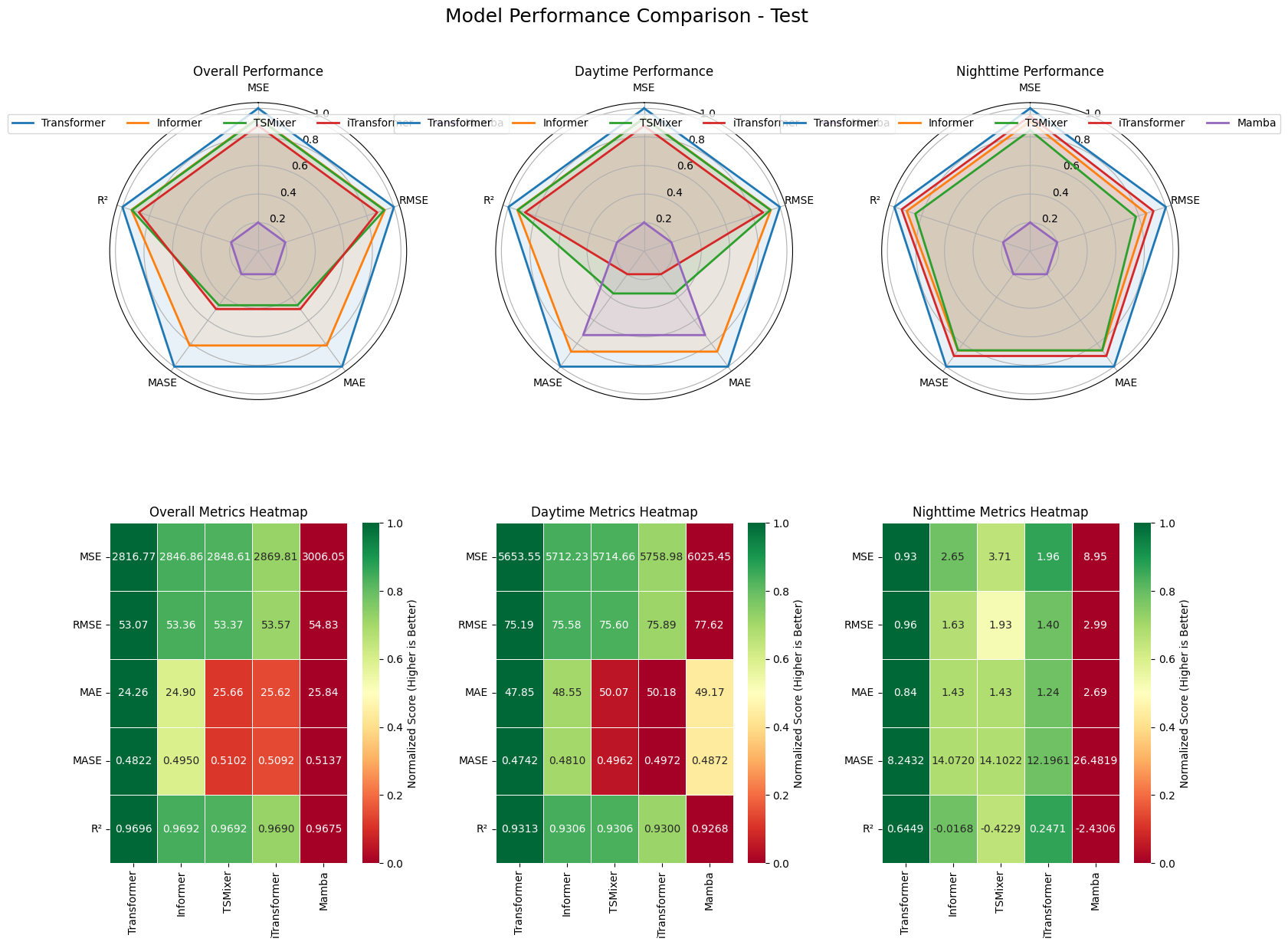}
    \caption{Comparing performance metrics of Advanced models across Overall, Daytime, and Nighttime conditions.}
    \label{fig:advanced_daytime_nighttime}
\end{figure}

Nighttime accuracy reflects a model's stability and ability to predict zero values without noise. The \textbf{Transformer} maintains its lead in this regime (Figure \ref{fig:advanced_daytime_nighttime}) with the lowest error metrics (MSE: 0.93) and the highest $R^2$ (0.6449). Notably, many other advanced models, including \textbf{Mamba}, perform poorly at night, exhibiting negative $R^2$ values (e.g., Mamba: -2.43).

For the basic models (Figure \ref{fig:basic_daytime_nighttime}), \textbf{TCN} dominates with an MSE of 0.67, being the only basic model to achieve a positive $R^2$ (0.7435). Other architectures like \textbf{LSTM} and \textbf{CNN-LSTM} show extremely negative $R^2$ values, suggesting they fail to fully suppress output signals when the ground truth is consistently zero.

\subsection{Prediction Pattern Visualization}
\begin{figure}[H]
    \centering
    \begin{subfigure}[b]{1.0\textwidth}
        \centering
        \includegraphics[width=\linewidth]{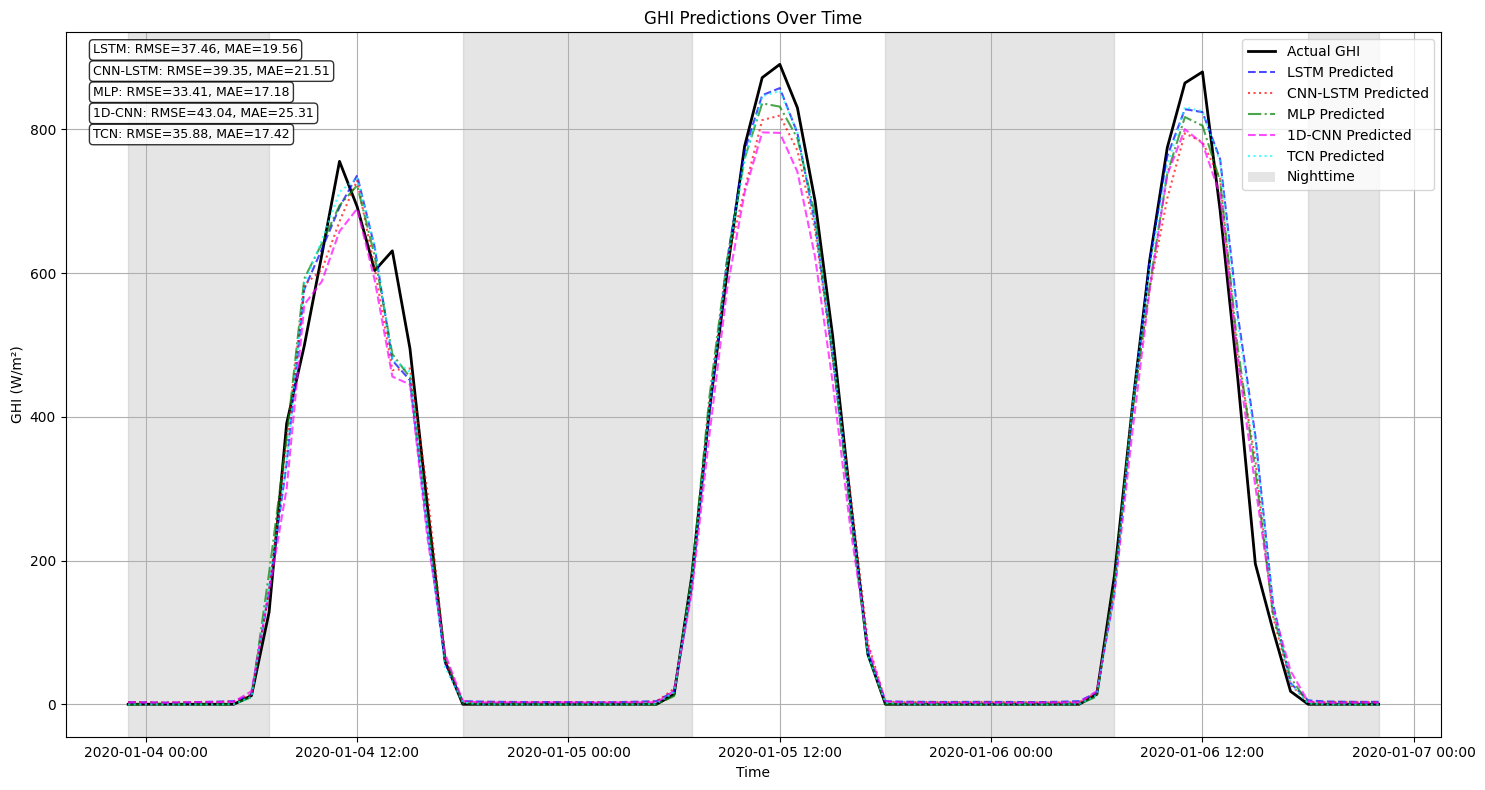}
        \caption{Basic Models Prediction}
    \end{subfigure}
    \hfill
    \begin{subfigure}[b]{1.0\textwidth}
        \centering
        \includegraphics[width=\linewidth]{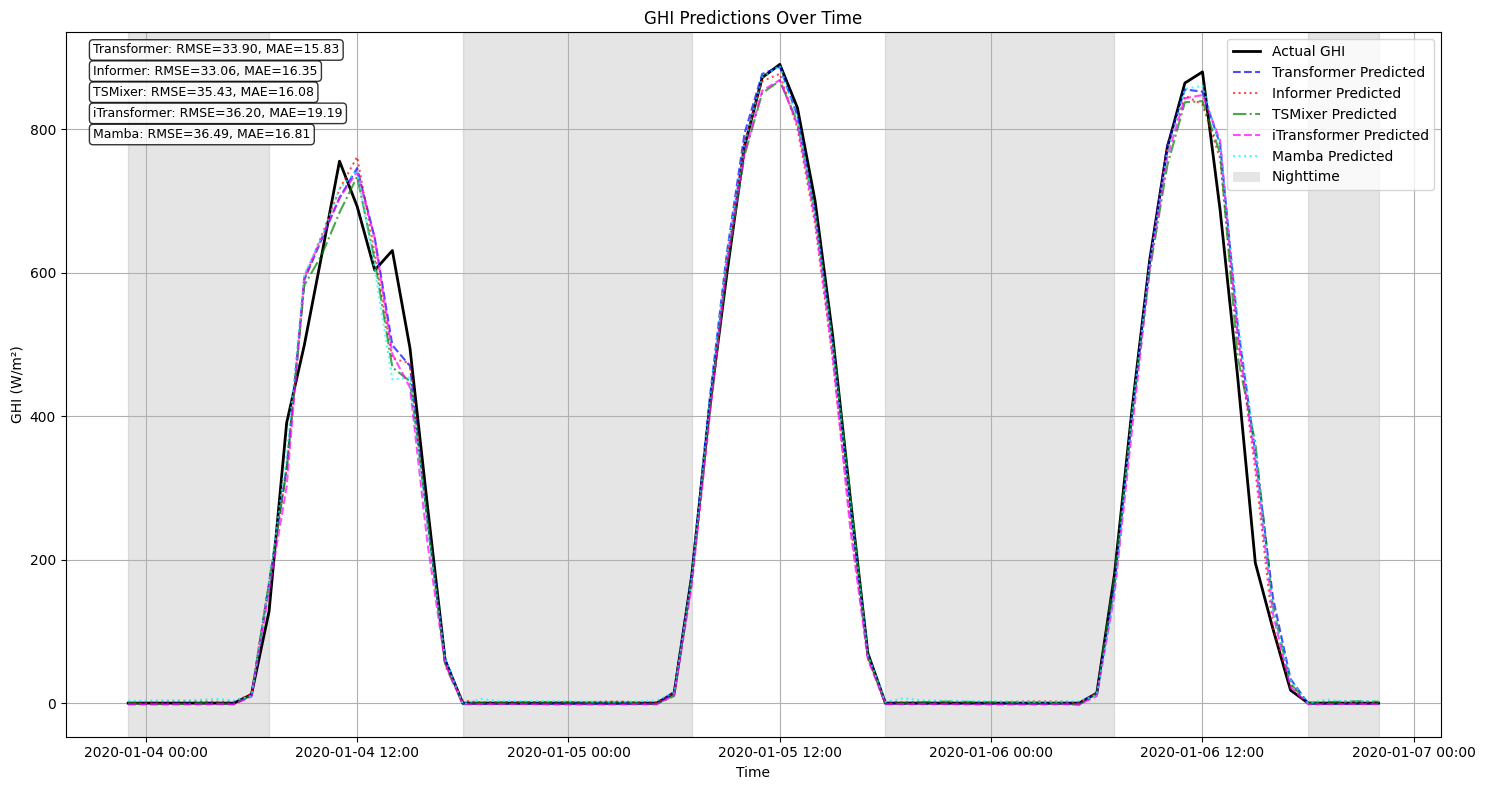}
        \caption{Advanced Models Prediction}
    \end{subfigure}
    \caption{Sample prediction over time of Basic and Advanced models on a few days of the test set.}
    \label{fig:prediction_sample}
\end{figure}

To qualitatively validate the models, we visualized their forecasts against ground truth GHI values over a sample period from the test set (January 4–7, 2020), as shown in Figure \ref{fig:prediction_sample}.

For the Basic Models (Figure \ref{fig:prediction_sample}a), \textbf{LSTM} and \textbf{TCN} track the actual GHI curve closely, capturing daily peaks effectively. \textbf{1D-CNN} shows larger deviations, often underestimating peak irradiance. All models correctly predict zero values during nighttime.

For the Advanced Models (Figure \ref{fig:prediction_sample}b), the predictions are generally tighter around the ground truth. The \textbf{Transformer} and \textbf{Informer} show excellent alignment with actual GHI peaks, particularly on clear days. \textbf{Mamba} exhibits slightly higher deviations but still maintains robust tracking of the diurnal cycle.

\subsection{Model Explainability}
\begin{wrapfigure}{r}{0.5\textwidth}
    \centering
    \includegraphics[width=0.8\textwidth]{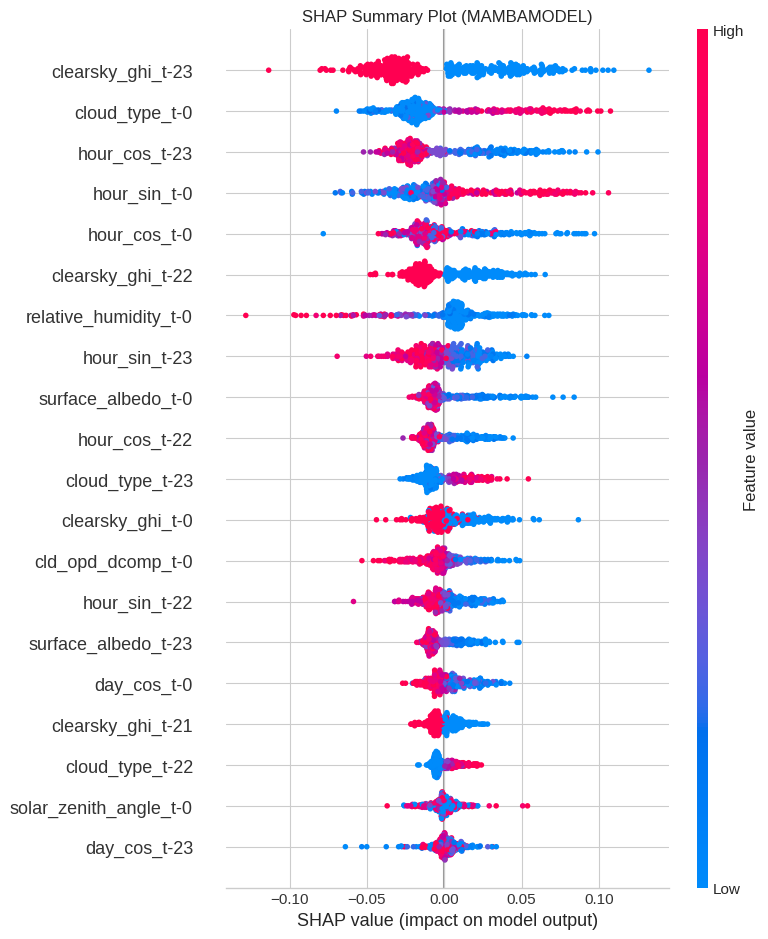}
    \caption{SHAP summary plot for Mamba model on 400 test daytime samples. Features are ranked by mean absolute SHAP value.}
    \label{fig:mamba_shap_summary}
\end{wrapfigure}

To interpret the decision-making process of the deep learning models, we employed SHapley Additive exPlanations (SHAP). Our analysis began with a global feature importance assessment for the Mamba model, followed by a comparative analysis of temporal attention mechanisms between state-space (Mamba) and attention-based (Transformer) architectures.

\subsubsection{Mamba Model Feature Analysis}

Figure \ref{fig:mamba_shap_summary} presents the SHAP summary plot for the Mamba model on 400 test daytime samples. The analysis reveals that temporal features dominate the model's explanatory power. Specifically, \texttt{clearsky\_ghi\_t-23}—the clear-sky GHI estimate from 23 time steps ago—exhibits the highest impact, with SHAP values reaching up to 0.10. This indicates a critical reliance on historical solar radiation patterns from the previous day. Similarly, \texttt{hour\_cos\_t-23} shows significant influence, underscoring the importance of cyclic hourly patterns.

Other notable contributors include \texttt{cloud\_type\_t-0}, \texttt{hour\_sin\_t-0}, and \texttt{surface\_albedo\_t-0}. For these features, high values (red dots) generally correlate with increased GHI predictions, while low values (blue dots) suggest reductions. Conversely, features such as \texttt{relative\_humidity\_t-0} and \texttt{solar\_zenith\_angle\_t-0} display SHAP values clustering near zero, indicating minimal contribution to the model's output.

\subsubsection{Comparative Temporal Analysis}
Further analysis revealed fundamentally different temporal attention mechanisms between the state-space (Mamba) and attention-based (Transformer) architectures.

\begin{figure}[H]
    \centering
    \begin{subfigure}[b]{0.9\textwidth}
        \centering
        \includegraphics[width=\linewidth]{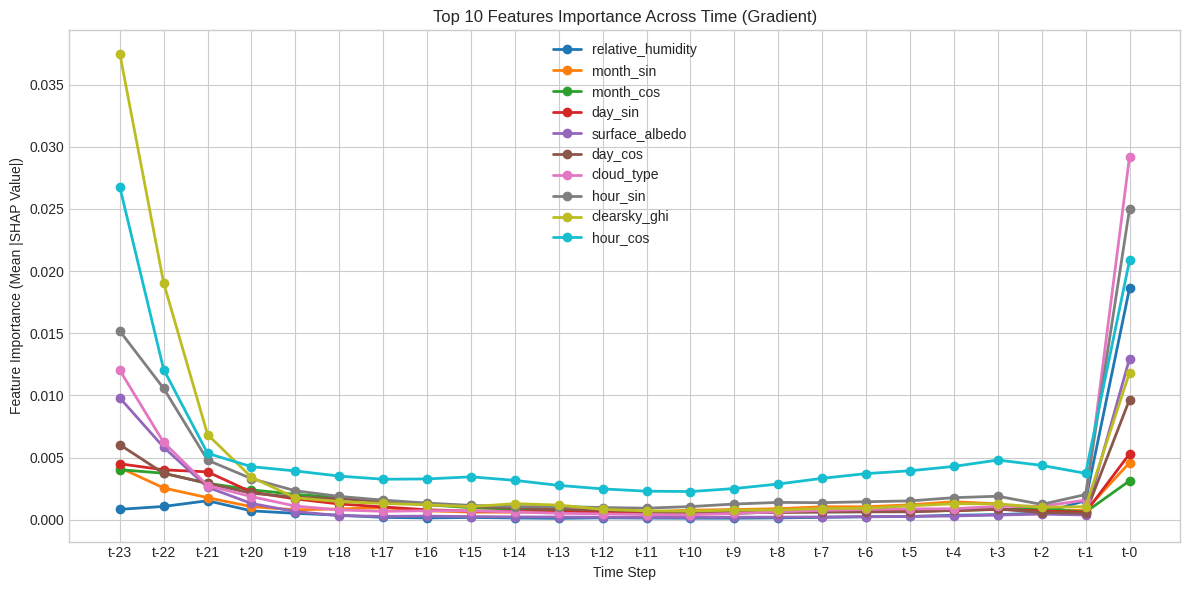}
        \caption{Mamba SHAP Analysis}
        \label{fig:mamba_shap_time}
    \end{subfigure}
    
    \vspace{0.3cm}
    
    \begin{subfigure}[b]{0.9\textwidth}
        \centering
        \includegraphics[width=\linewidth]{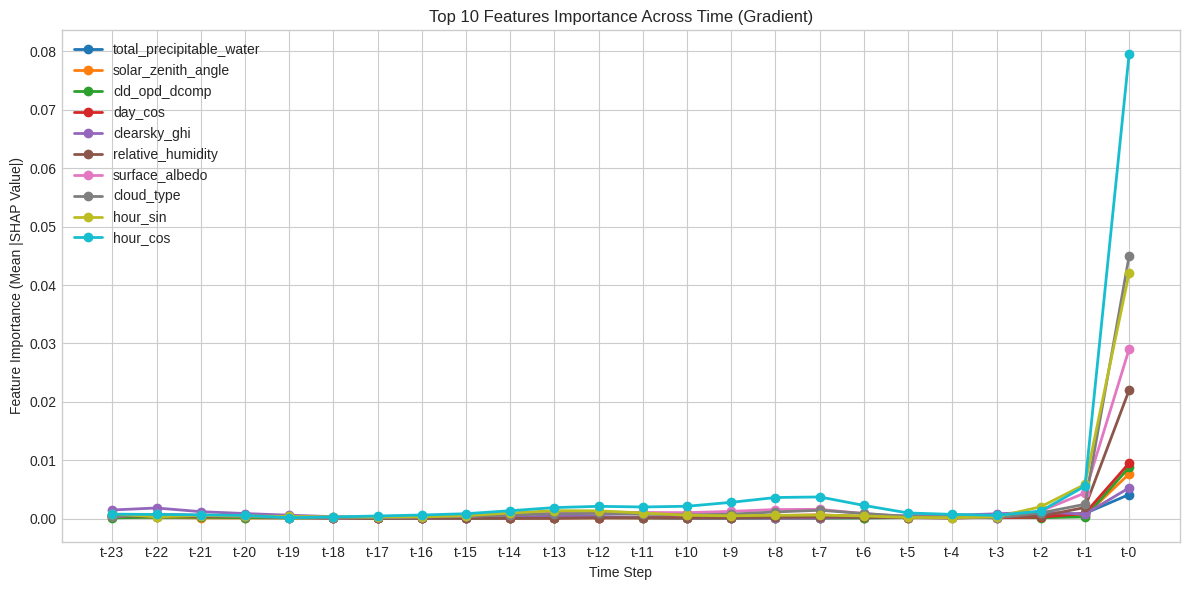}
        \caption{Transformer SHAP Analysis}
        \label{fig:transformer_shap_time}
    \end{subfigure}
    \caption{SHAP Top 10 Feature Importance analysis over timesteps comparing (a) Mamba and (b) Transformer models.}
    \label{fig:shap_comparison}
\end{figure}

As shown in Figure \ref{fig:shap_comparison}(a), the \textbf{Mamba} model exhibits a distinct \textbf{U-shaped temporal pattern}. It places significant weight on both the immediate past ($t-0$) and the distant history ($t-23$), effectively capturing the 24-hour daily cycle. The most critical features for Mamba are \texttt{clearsky\_ghi} and \texttt{hour\_cos} at $t-23$, reinforcing its reliance on historical periodic trends to infer current irradiance.

In contrast, the \textbf{Transformer} model (Figure \ref{fig:shap_comparison}(b)) demonstrates an extreme \textbf{recency bias}. Nearly all feature importance is concentrated at the most recent time step ($t-0$), with virtually negligible values across all prior time steps from $t-23$ to $t-1$. The dominant features at $t-0$ are \texttt{hour\_cos}, \texttt{cloud\_type}, and \texttt{hour\_sin}.

This comparison suggests that the Transformer relies almost exclusively on the immediate state—current time of day and cloud conditions—to make its prediction, largely ignoring the historical sequence. Mamba, conversely, leverages its state-space memory to synthesize information from the previous day's cycle ($t-23$), indicating a more balanced but structurally different approach to temporal reasoning.

\section{Model Compression}
\label{sec:compression}

To facilitate deployment in resource-constrained environments, we applied three model compression techniques to the best-performing \textbf{Transformer} model: Quantization, Structured Pruning, and Knowledge Distillation. The objective was to reduce model size and inference latency while maintaining predictive accuracy.

Table \ref{tab:compression_results_comprehensive} presents the comprehensive results, comparing each technique against its specific baseline environment (e.g., ONNX Runtime for quantization vs. standard PyTorch for pruning/distillation).

\begin{table}[htbp]
\centering
\caption{Comprehensive Comparison of Model Compression Techniques}
\label{tab:compression_results_comprehensive}
\resizebox{\textwidth}{!}{%
\begin{tabular}{lcccc}
\hline
\textbf{Method} & \textbf{Model Size (MB)} & \textbf{Size Reduction (\%)} & \textbf{Latency (ms)} & \textbf{MAE (W/m$^2$)} \\ \hline
\multicolumn{5}{c}{\textit{\textbf{Quantization (CPU - ONNX Runtime)}}} \\
Baseline (ONNX FP32) & 1.21 & - & 547.61 & 24.26 \\
Int8 Quantization & 0.44 & \textbf{64.0\%} & 519.88 & 25.24 \\
Int4 Quantization & \textbf{0.11} & \textbf{91.0\%} & 589.69 & 25.60 \\ \hline
\multicolumn{5}{c}{\textit{\textbf{Quantization (GPU - ONNX Runtime)}}} \\
Baseline (ONNX FP32) & 1.21 & - & 41.52 & 24.26 \\
FP16 Quantization & 0.65 & 46.0\% & \textbf{22.02} & \textbf{24.25} \\ \hline
\multicolumn{5}{c}{\textit{\textbf{Structured Pruning (PyTorch)}}} \\
Baseline (Original) & 1.07 & - & 3792.13 & 24.26 \\
Pruning (10\% Sparsity) & 1.07 & 0.0\% & 3808.80 & 40.66 \\
Pruning (30\% Sparsity) & 1.07 & 0.0\% & 3849.29 & 106.17 \\
Pruning (50\% Sparsity) & 1.07 & 0.0\% & 3857.31 & 176.21 \\ \hline
\multicolumn{5}{c}{\textit{\textbf{Knowledge Distillation (PyTorch)}}} \\
Teacher (Baseline) & 1.07 & - & 3792.13 & 24.26 \\
Student (From Scratch) & 0.82 & 23.5\% & 3069.28 & 25.35 \\
\textbf{Student (Distilled)} & 0.82 & 23.5\% & 3081.46 & \textbf{23.78} \\ \hline
\end{tabular}%
}
\end{table}

\subsection{Quantization}
Quantization was implemented using the ONNX runtime to reduce the precision of weights and activations. We evaluated dynamic quantization for CPU (Int8, Int4) and mixed-precision for GPU (FP16).

\textbf{GPU FP16 Quantization} yielded the most practical improvements for high-performance deployment. It reduced inference latency by approximately 47\% (from 41.52 ms to 22.02 ms) while maintaining near-identical accuracy to the baseline (MAE: 24.25 vs 24.26 W/m$^2$).

For CPU deployment, \textbf{Int8 Quantization} effectively reduced the model size by 64\% (from 1.21 MB to 0.44 MB) with only a minor degradation in accuracy (MAE increased by $\approx$1 W/m$^2$). However, aggressive \textbf{Int4 Quantization}, despite achieving a 91\% size reduction, resulted in higher latency and further accuracy loss, making it less suitable for this forecasting task.

\subsection{Structured Pruning}
We applied structured pruning to the Transformer's linear layers at sparsity levels of 10\%, 30\%, and 50\%. As shown in Table \ref{tab:compression_results_comprehensive}, this technique proved \textbf{ineffective} for our specific implementation. 

Due to standard PyTorch storage mechanisms, the physical model size remained unchanged (1.07 MB) despite the theoretical sparsity. More critically, the predictive performance degraded severely even at low sparsity levels; at 50\% pruning, the MAE spiked to 176.21 W/m$^2$. This suggests that the Transformer's dense connections are highly sensitive to channel removal in this time-series context.

\subsection{Knowledge Distillation}
Knowledge Distillation proved to be the most robust compression strategy. We trained a smaller "Student" model to mimic the outputs of the larger "Teacher" Transformer.

The \textbf{Distilled Student} model achieved a unique balance of speed and accuracy. It reduced model size by 23.5\% and latency by approximately 19\% compared to the Teacher. Remarkably, it outperformed the Teacher model in accuracy (MAE: 23.78 vs 24.26 W/m$^2$).

\begin{figure}[htbp]
    \centering
    \includegraphics[width=\linewidth]{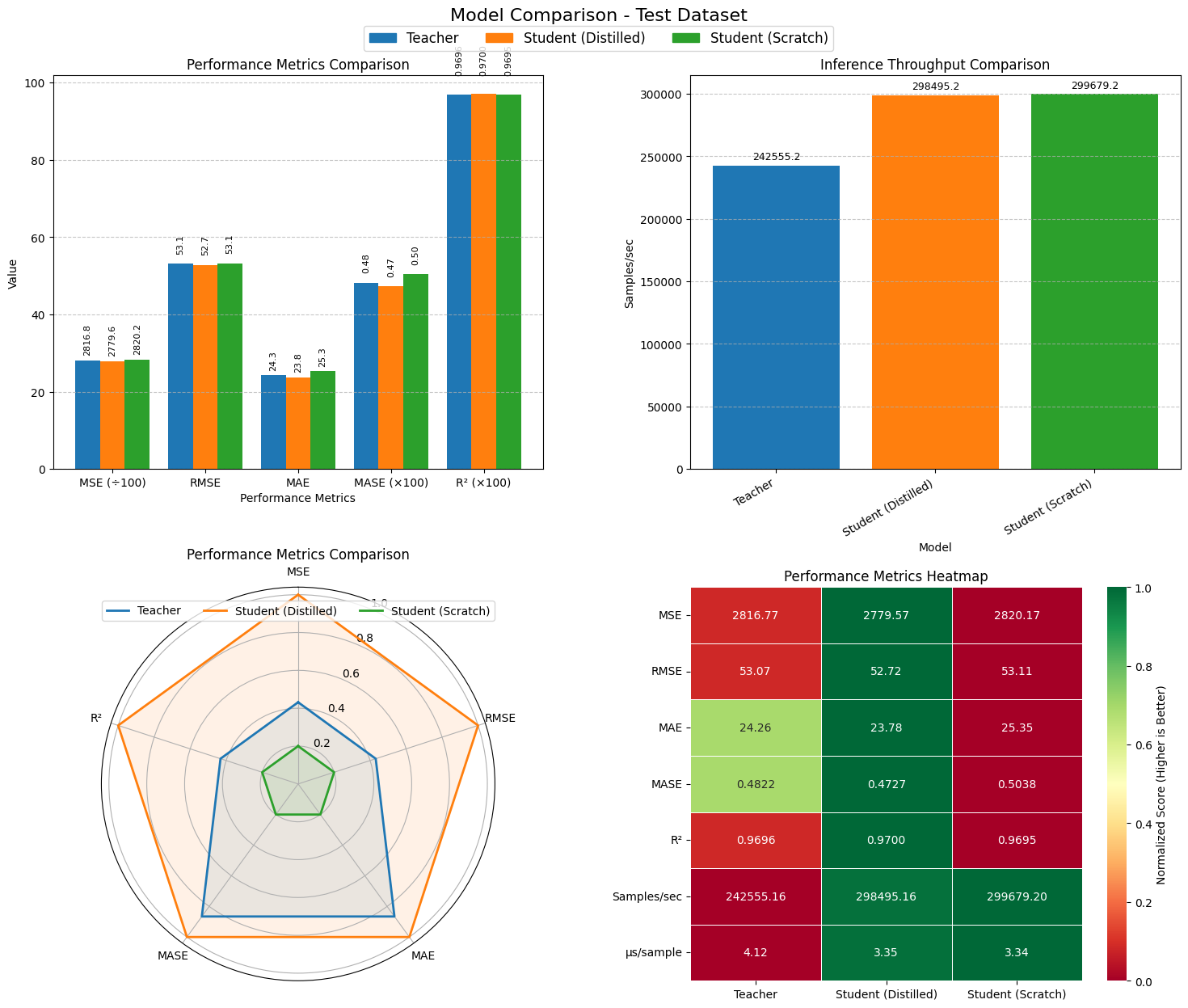}
    \caption{Model Comparison between Transformer models: Teacher, Student with Distillation loss, and Student trained from scratch.}
    \label{fig:distillation_charts}
\end{figure}

Figure \ref{fig:distillation_charts} highlights that the Distilled Student also outperformed an identical student model trained from scratch (MAE: 25.35 W/m$^2$). This confirms that the distillation process acts as a powerful regularizer, enabling the smaller network to learn more generalizable patterns than it could discover independently.

\newpage
\section{Conclusion}
\label{sec:conclusion}

This research has presented a comprehensive evaluation of deep learning architectures for short-term Global Horizontal Irradiance (GHI) forecasting in Ho Chi Minh City, utilising high-resolution satellite data from the NSRDB. By comparing ten distinct models ranging from basic sequence networks to state-of-the-art Transformers and State Space Models (Mamba), this study offers significant insights into the trade-offs between predictive accuracy, computational efficiency, and model interpretability.

\subsection{Key Findings}

\textbf{Model Performance and Robustness:}
The \textbf{Transformer} architecture emerged as the superior model, achieving the lowest overall errors (MSE: 2816.77, MAE: 24.26) and demonstrating remarkable stability across both daytime and nighttime regimes. Among the basic architectures, the \textbf{Temporal Convolutional Network (TCN)} proved to be a highly competitive alternative, matching the performance of more complex advanced models while offering lower structural complexity. Our temporal analysis highlighted a critical distinction: while the Transformer and TCN remained robust during nighttime (predicting near-zero values accurately), other advanced architectures like Mamba and Informer struggled with noise suppression in zero-irradiance conditions.

\textbf{Interpretability and Temporal Attention:}
A novel contribution of this work is the comparative SHAP analysis between attention-based and state-space architectures. We identified fundamentally different mechanisms for temporal reasoning. The \textbf{Mamba} model exhibited a "U-shaped" attention pattern, explicitly leveraging historical context from the previous day ($t-23$) to inform predictions. in contrast, the \textbf{Transformer} demonstrated a strong "recency bias," relying almost exclusively on immediate atmospheric inputs ($t-0$). This suggests that while Mamba attempts to model the physics of periodicity, the Transformer excels by optimising for the immediate state of the atmosphere.

\textbf{Efficiency and Deployment:}
Our investigation into model compression revealed that \textbf{Knowledge Distillation} is the most effective strategy for deployment. The distilled student model achieved a rare "win-win" outcome: reducing model size by 23.5\% and latency by 19\% whilst simultaneously outperforming the teacher model in accuracy (MAE: 23.78 W/m$^2$). This validates the potential for deploying high-performance AI on resource-constrained edge devices at solar farms. Conversely, structured pruning proved ineffective for this time-series application, severely degrading performance.

\textbf{Mamba vs Transformer Efficiency:}
Despite Mamba's theoretical linear scaling $O(L)$, it demonstrated lower inference throughput (193,084 samples/sec) compared to the quadratic $O(L^2)$ Transformer (239,871 samples/sec).
\textit{*Hypothesis:} We attribute this discrepancy to the sequence length regime inherent to this task. The input window of $L=24$ is relatively short. At this scale, the quadratic cost of self-attention is negligible, allowing the Transformer to fully exploit massive GPU parallelisation. Conversely, Mamba's sequential scan mechanism—while efficient for long sequences ($L > 1000$)—incurs fixed kernel launch overheads and memory access costs that outweigh its complexity benefits on short sequences. This finding suggests that while State Space Models are promising for long-term forecasting, highly parallelized Transformers remain superior for short-context operational horizons.

\subsection{Broader Impact}

These findings directly support UN Sustainable Development Goals 7 (Affordable and Clean Energy) and 13 (Climate Action). The developed Transformer and TCN models provide the precise forecasting required to stabilise power grids against solar variability, facilitating higher penetrations of renewable energy. Furthermore, the success of the distilled models demonstrates that AI-driven sustainability solutions need not be computationally exorbitant, aligning with the principle of "Green AI".

\subsection{Future Work}

To further advance the scalability and operational utility of this framework, future research will focus on three key directions:

\begin{itemize}
    \item \textbf{Multi-region Model Generalization:} 
    While this study focused on the tropical climate of Ho Chi Minh City, solar irradiance patterns vary significantly across latitudes and terrains. Future experiments will evaluate model transferability across diverse climatic zones—such as the arid South Central Coast and the monsoonal Northern region—to develop a universal "foundation model" for national-scale solar forecasting.

    \item \textbf{Long-term GHI Forecasting:} 
    Operational grid scheduling requires planning horizons beyond the current 1-hour window. We aim to extend the forecasting horizon to 24 and 48 hours to support day-ahead market bidding and battery storage optimization. This will likely require revisiting the \textbf{Mamba} architecture, as its linear scaling properties ($O(L)$) may prove superior to Transformers when processing the longer sequence histories required for day-ahead predictions.

    \item \textbf{Extended Weather Variable Integration:} 
    To refine predictive precision, future iterations will enrich the input feature space beyond standard meteorological variables. We propose integrating high-dimensional dynamic data, such as real-time Aerosol Optical Depth (AOD) to account for urban pollution, and raw satellite imagery channels to capture cloud texture and velocity directly, rather than relying solely on tabular derivatives.
\end{itemize}
\section*{Acknowledgements}

The author gratefully acknowledges Dr. Erick Giovani Sperandio Nascimento, module leader of EEEM073, for his academic guidance and constructive feedback throughout this module. Sincere thanks are also extended to the National Renewable Energy Laboratory (NREL) for providing access to the National Solar Radiation Database (NSRDB), which was instrumental to this study. Finally, the author appreciates the support of the University of Surrey’s High Performance Computing facilities for enabling the deep learning experiments presented in this report.

\bibliographystyle{elsarticle-num-names}
\bibliography{references}

\clearpage
\appendix
\section{Full List of Features in NSRDB Himawari 7 Data}
\label{appendix:nsrdb_features}

The following tables detail the complete set of variables available in the NSRDB Himawari 7 satellite-derived dataset used for this study. The features are categorised into meteorological, solar/atmospheric, target variables, and metadata. Range values and means were computed based on the filtered dataset for Ho Chi Minh City (2011--2020).

\begin{table}[H]
\centering
\caption{Meteorological Variables}
\label{tab:met_features}
\resizebox{\textwidth}{!}{%
\begin{tabular}{lllccc}
\hline
\textbf{Variable} & \textbf{Data Type} & \textbf{Description} & \textbf{Unit} & \textbf{Range} & \textbf{Mean} \\ \hline
Air Temperature & int16 & Surface air temperature & K & 222 -- 289 & 262.13 \\
Dew Point & int16 & Temperature at which air saturates & K & 198 -- 219 & 210.2 \\
Relative Humidity & uint16 & Percentage of moisture relative to saturation & \% & 62.33 -- 88.57 & 73.53 \\
Wind Speed & uint16 & Surface wind speed & m/s & 0.2 -- 2.9 & 1.72 \\
Wind Direction & uint16 & Direction of wind origin & degrees & 18 -- 245 & 82.51 \\
Surface Pressure & uint16 & Atmospheric pressure at surface & hPa & 1007 -- 1011 & 1009.67 \\
Total Precipitable Water & uint8 & Total column water vapor & mm & 40 -- 48 & 44.9 \\ \hline
\end{tabular}%
}
\end{table}

\begin{table}[H]
\centering
\caption{Solar and Atmospheric Variables}
\label{tab:solar_features}
\resizebox{\textwidth}{!}{%
\begin{tabular}{lllccc}
\hline
\textbf{Variable} & \textbf{Data Type} & \textbf{Description} & \textbf{Unit} & \textbf{Range} & \textbf{Mean} \\ \hline
Solar Zenith Angle & uint16 & Angle between sun and vertical & degrees & 33.98 -- 80.23 & 53.98 \\
Clearsky GHI & uint16 & Clear-sky global horizontal irradiance & W/m$^2$ & 109 -- 829 & 530.66 \\
Clearsky DHI & uint16 & Clear-sky diffuse horizontal irradiance & W/m$^2$ & 63 -- 248 & 162.85 \\
Clearsky DNI & uint16 & Clear-sky direct normal irradiance & W/m$^2$ & 263 -- 769 & 600.49 \\
Aerosol Optical Depth & uint16 & Aerosol extinction in atmosphere & unitless & 0.22 -- 0.43 & 0.30 \\
Surface Albedo & uint8 & Fraction of reflected solar radiation & unitless & 0.07 -- 0.18 & 0.14 \\
Ozone & uint16 & Total column ozone & DU & 238 -- 240 & 239.32 \\
Cloud Type & int8 & Categorical cloud presence/type & int & 0 -- 12 & - \\
Cloud Optical Depth & uint16 & Cloud optical thickness & unitless & 0 -- 80 & 1.30 \\
Cloud Effective Radius & uint16 & Cloud particle effective radius & $\mu$m & 0 -- 18.9 & 1.21 \\
Cloud Pressure & uint16 & Pressure at cloud top & hPa & 0 -- 916 & 85.85 \\
Single Scattering Albedo & uint8 & Aerosol scattering-to-extinction ratio & unitless & 0.93 -- 0.96 & 0.94 \\
Asymmetry Parameter & int8 & Aerosol scattering directionality & unitless & - & - \\
Alpha & uint8 & Ångström exponent for aerosol size & unitless & 1.08 -- 1.53 & 1.34 \\ \hline
\end{tabular}%
}
\end{table}

\begin{table}[H]
\centering
\caption{Target Variables and Quality Flags}
\label{tab:target_features}
\resizebox{\textwidth}{!}{%
\begin{tabular}{lllccc}
\hline
\textbf{Variable} & \textbf{Data Type} & \textbf{Description} & \textbf{Unit} & \textbf{Range} & \textbf{Mean} \\ \hline
\textbf{GHI (Target)} & \textbf{uint16} & \textbf{Global Horizontal Irradiance} & \textbf{W/m$^2$} & \textbf{9 -- 829} & \textbf{510.79} \\
DHI & uint16 & Diffuse Horizontal Irradiance & W/m$^2$ & 9 -- 560 & 173.38 \\
DNI & uint16 & Direct Normal Irradiance & W/m$^2$ & 0 -- 769 & 539.08 \\ \hline
Fill Flag & uint8 & Indicates missing/filled data & int & 0 -- 3 & 0.112 \\
Cloud Fill Flag & uint8 & Indicates cloud data filling status & int & 0 -- 3 & 0.112 \\ \hline
\end{tabular}%
}
\end{table}

\begin{table}[H]
\centering
\caption{Metadata}
\label{tab:metadata}
\resizebox{\textwidth}{!}{%
\begin{tabular}{llll}
\hline
\textbf{Variable} & \textbf{Data Type} & \textbf{Description} & \textbf{Example/Range} \\ \hline
Latitude & float32 & Grid cell latitude & 10.37 -- 10.53$^{\circ}$ \\
Longitude & float32 & Grid cell longitude & 106.42 -- 106.58$^{\circ}$ \\
Elevation & int16 & Surface elevation & 0 -- 4 m \\
Timezone & int16 & Time zone offset & +7 (UTC) \\
Country & string & Country name & Vietnam \\
State & string & State name & Ho Chi Minh \\
County & string & County name & e.g., Binh Chanh, Cu Chi \\ \hline
\end{tabular}%
}
\end{table}

\section{Transformer Model - Training history, Test Prediction Distribution}
\label{appendix:transformer_analysis}

\begin{figure}[H]
    \centering
    \includegraphics[width=\textwidth]{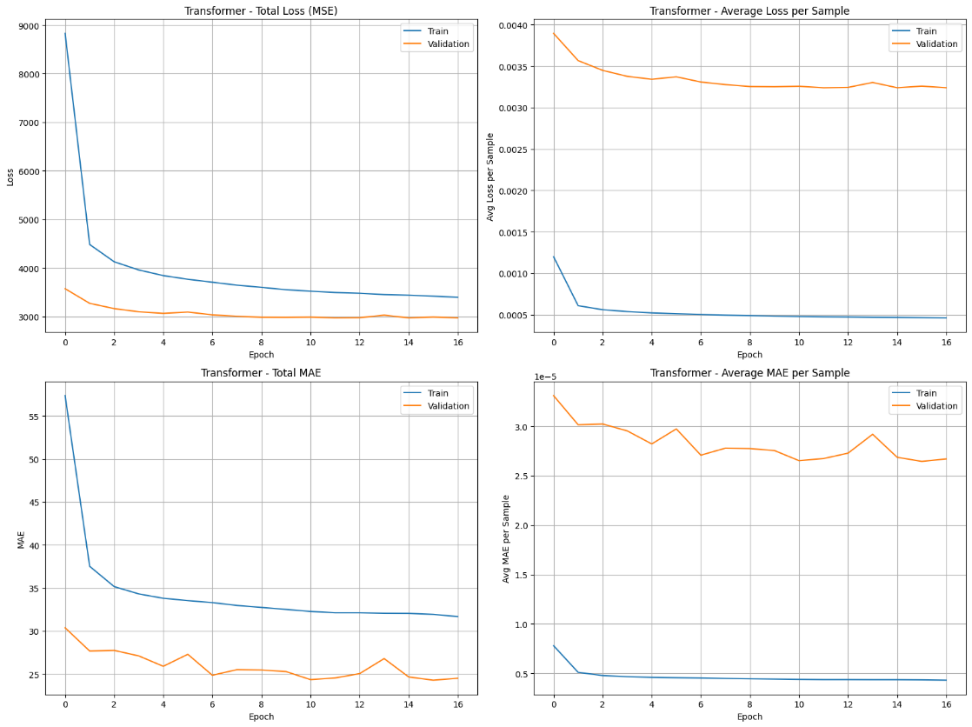} 
    \caption{Training and Validation learning curve of Transformer model.}
    \label{fig:transformer_learning_curve}
\end{figure}

In Figure \ref{fig:transformer_learning_curve}, the Transformer model's training and validation chart reveals effective learning over 17 epochs, with Total Loss (MSE) and Total MAE (Bottom Left) decreasing sharply for both training (blue) and validation (orange) sets, stabilizing below 1000 W/m$^2$ and 30 W/m$^2$ respectively, indicating good convergence.

\begin{figure}[H]
    \centering
    \includegraphics[width=0.8\textwidth]{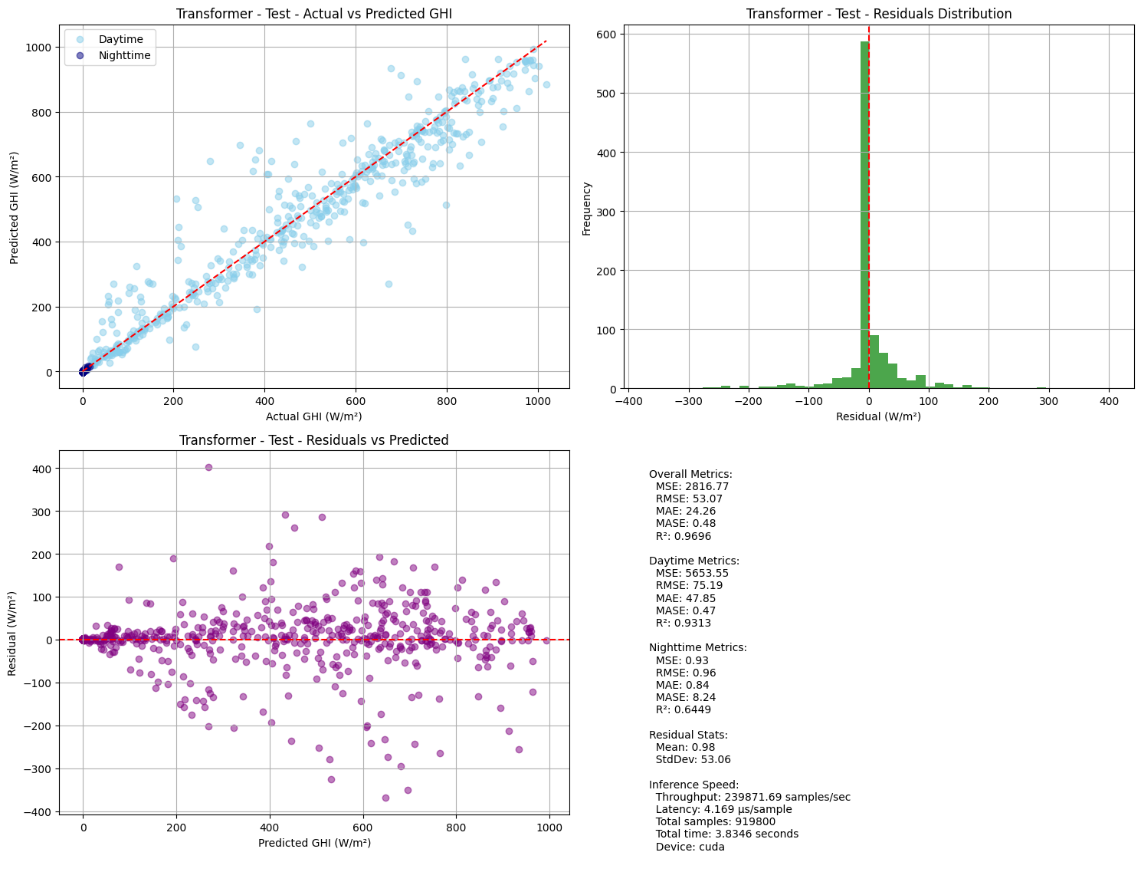} 
    \caption{Prediction/Residuals Distribution and Trend of Transformer model on Test set.}
    \label{fig:transformer_residuals}
\end{figure}

In Figure \ref{fig:transformer_residuals}, the Transformer model's Actual vs. Predicted GHI plot (Top Left) shows good alignment along the diagonal ($R^2=0.9649$), but scatters at higher GHI values (400–800 W/m$^2$), indicating under- or over-predictions during peak irradiance. The Residuals Distribution (Top Right) is centred near 0 (Mean 0.58 W/m$^2$) with a StdDev of 53.06 W/m$^2$, but exhibits slight negative skew and outliers up to 300 W/m$^2$, reflecting occasional large errors. The Residuals vs. Predicted GHI plot (Bottom Left) reveals heteroscedasticity, with residuals spreading wider (up to ±300 W/m$^2$) at higher predicted GHI, highlighting increased uncertainty during daytime high-irradiance periods (daytime RMSE 75.19 W/m$^2$).

\section{Mamba Model Feature Sensitivity}
\label{appendix:mamba_sensitivity}

This section visualises the global feature sensitivity of the Mamba model, generated by the \texttt{Sensitive Analyzer}. This analysis perturbs each feature by 10\% to measure the mean absolute impact on the model's output, offering a holistic view of which features drive the model's decision-making process.

\vspace{0.5cm} 

\begin{figure}[H]
    \centering
    \hspace*{-2.5cm}%
    \begin{minipage}{0.65\textwidth}
        \centering
        \includegraphics[width=\linewidth]{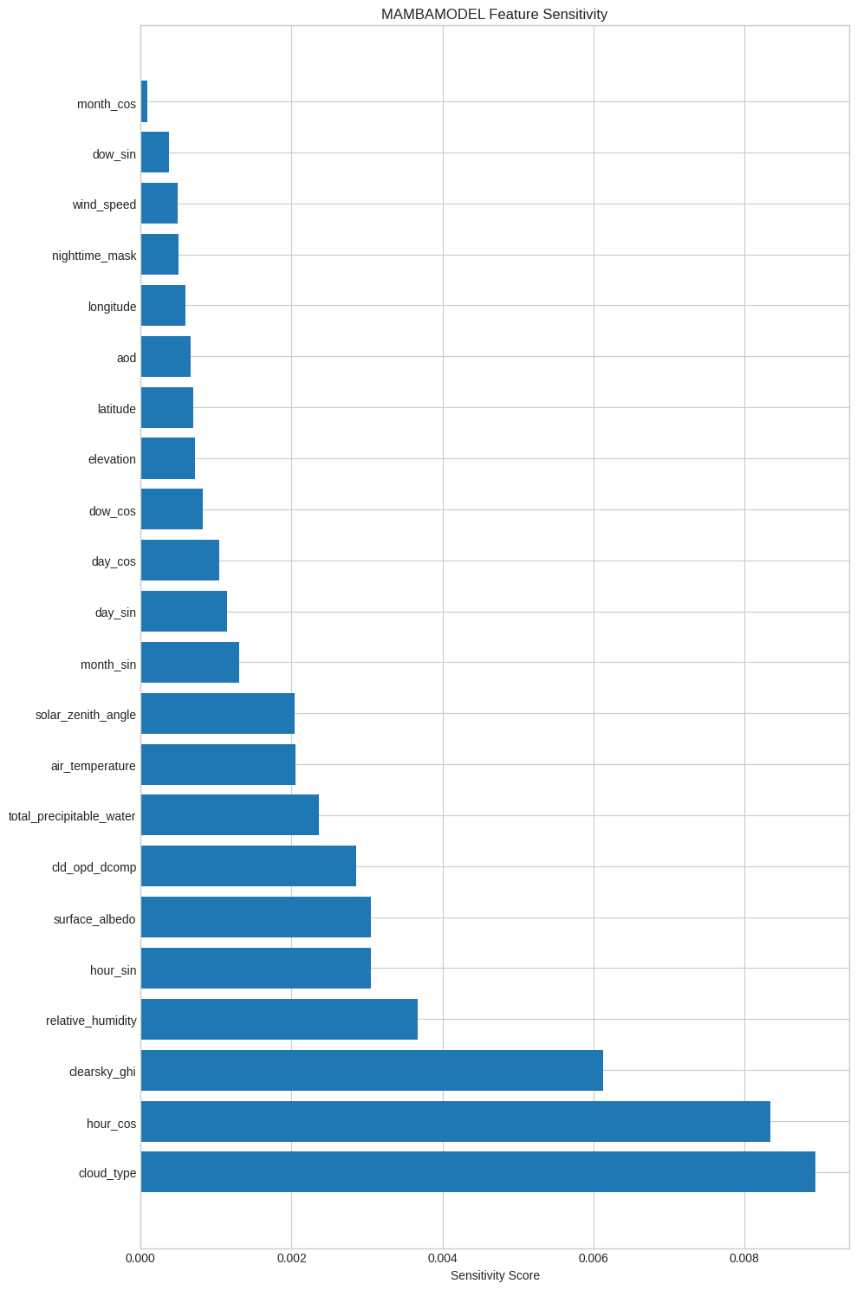}
        \caption{The "MAMBA MODEL Feature Sensitivity" plot, generated by the Sensitive Analyzer}
        \label{fig:mamba_sensitivity}
    \end{minipage}%
    \hfill
    \begin{minipage}{0.5\textwidth}
        As shown in Figure \ref{fig:mamba_sensitivity}, the Mamba model is most sensitive to \texttt{cloud\_type} (sensitivity $\approx$ 0.008) and \texttt{hour\_cos} ($\approx$ 0.007). This emphasises the model's reliance on real-time weather conditions and hourly temporal patterns for accurate forecasting. \texttt{clearsky\_ghi} also plays a significant role ($\approx$ 0.0045), providing a physical baseline for irradiance.
        Notably, the \texttt{nighttime\_mask} shows very low sensitivity, which aligns with the analysis design that focused exclusively on daytime samples where GHI $>0$. In contrast, features such as \texttt{month\_cos}, \texttt{dow\_sin}, and \texttt{wind\_speed} exhibit minimal impact on the model's output. This suggests that these variables could potentially be pruned to streamline the model without compromising accuracy, highlighting the Mamba architecture's ability to filter out less relevant static or low-frequency signals in favour of dynamic, time-critical inputs.

    \end{minipage}
\end{figure}

\section{SHAP waterfall visualisation for high and mid-range prediction instance}
\label{appendix:shap_waterfall}

This section visualizes the feature attribution for specific individual predictions using SHAP waterfall plots on the Mamba model.

\begin{figure}[htbp]
    \centering
    \begin{subfigure}[t]{0.48\textwidth}
        \centering
        \includegraphics[width=\textwidth]{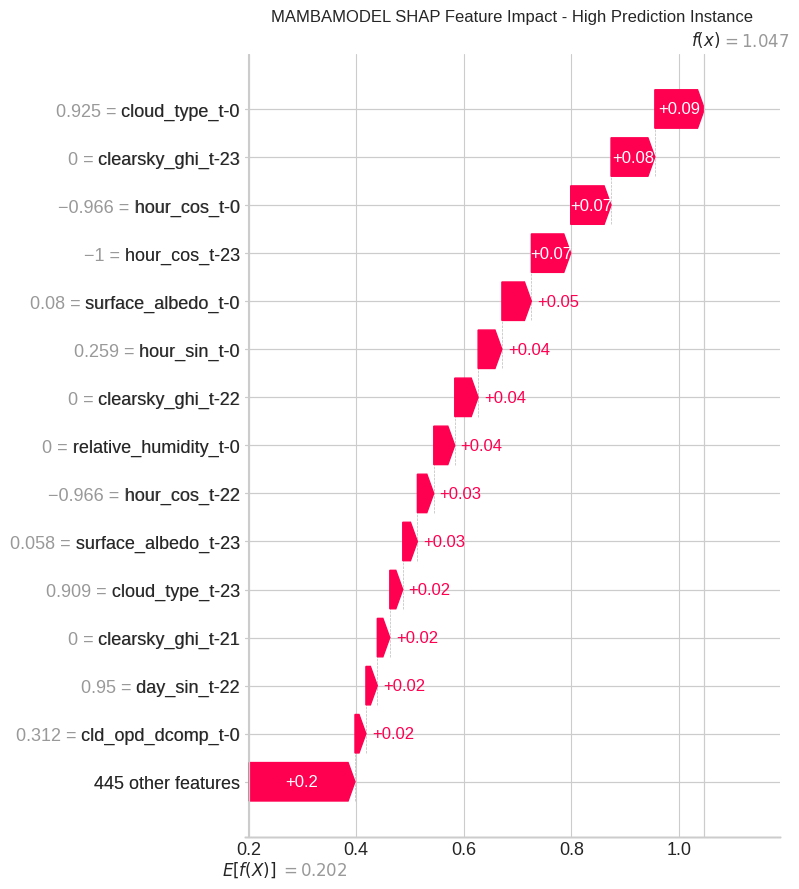}
        \caption{SHAP waterfall for high prediction.}
        \label{fig:mamba_waterfall_high}
    \end{subfigure}
    \hfill
    \begin{subfigure}[t]{0.48\textwidth}
        \centering
        \includegraphics[width=\textwidth]{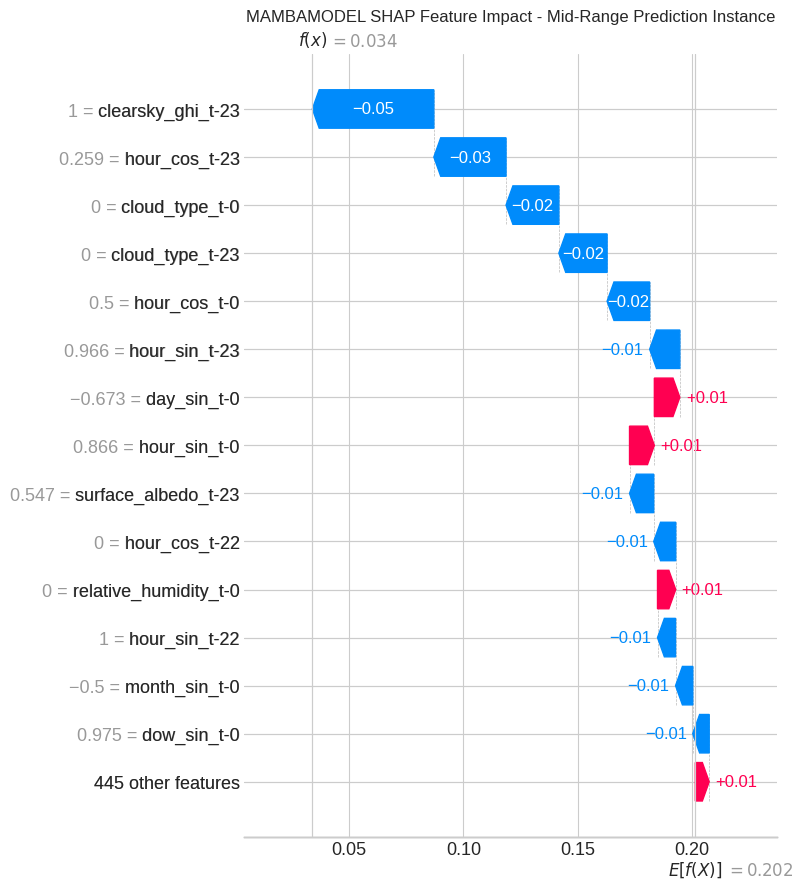}
        \caption{SHAP waterfall for mid-range value prediction.}
        \label{fig:mamba_waterfall_mid}
    \end{subfigure}
    \caption{SHAP waterfall plots for Mamba model predictions.}
    \label{fig:mamba_waterfall}
\end{figure}

Figure \ref{fig:mamba_waterfall} illustrates the feature contributions for two distinct instances:
\begin{itemize}
    \item \textbf{High Prediction Instance ($f(x) = 1.047$):} The model starts from a base value $E[f(x)] = 0.202$. Positive contributions dominate, driven by \texttt{cloud\_type} at $t-0$ (+0.09) and \texttt{clearsky\_ghi} at $t-23$ (+0.08). The U-shaped temporal pattern is evident, with significant contributions from both immediate ($t-0$) and 24-hour lagged ($t-23$) features.
    \item \textbf{Mid-Range Prediction Instance ($f(x) = 0.034$):} Here, the prediction is pulled below the base value by negative contributions. \texttt{clearsky\_ghi} at $t-23$ has the largest negative impact (-0.05), followed by \texttt{hour\_cos} at $t-23$ (-0.03). This confirms that the model relies on the same key features (lagged and immediate) regardless of whether the outcome is high or low irradiance.
\end{itemize}

\end{document}